\documentclass[11pt]{article}

\usepackage[preprint]{acl}

\usepackage{times}
\usepackage{latexsym}

\usepackage[T1]{fontenc}

\usepackage[utf8]{inputenc}

\usepackage{microtype}

\usepackage{inconsolata}

\usepackage{graphicx}

\usepackage{algorithm}
\usepackage{algpseudocode}
\usepackage{amsmath}
\usepackage{amsfonts}
\usepackage{amsthm}
\usepackage{amssymb}
\usepackage{booktabs}
\usepackage{multirow}
\usepackage{multicol}
\usepackage{makecell}
\usepackage{enumerate}
\usepackage{enumitem}
\usepackage{caption}
\usepackage{algpseudocode}
\usepackage{xcolor}
\usepackage{float}
\usepackage{graphicx}
\usepackage{color}
\usepackage{xspace}
\usepackage{tikz}
\usetikzlibrary{arrows.meta,positioning,decorations.pathreplacing}
\usepackage{subcaption}
\usepackage{placeins}
\usepackage[table]{xcolor}
\usepackage[most]{tcolorbox}
\tcbset{
  colback=gray!5,
  colframe=gray!70,
  boxrule=0.5pt,
  arc=6.5pt,
  left=6pt,right=6pt,top=4pt,bottom=4pt,
}
\newtcolorbox{highlightbox}{}

\usepackage{booktabs}
\usepackage{tabularx}
\usepackage{array}

 \newcommand{\model}{\texttt{ITP}\xspace}
\newcommand{\gr}{\cellcolor{gray!15}}
\newcommand{\bl}{\cellcolor{blue!10}}

\usepackage{todonotes}

\newcommand{\qwenicon}{\raisebox{-0.2\height}{\includegraphics[height=2.4ex]{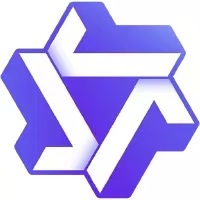}}}
\newcommand{\llamaicon}{\raisebox{-0.2\height}{\includegraphics[height=2.0ex]{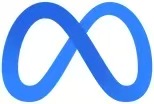}}}
\usepackage{xcolor}
\usepackage{pifont}

\newcommand{\cmark}{\ding{51}}
\newcommand{\xmark}{\ding{55}}
\usepackage{hyperref}
\usepackage{fontawesome5}
\hypersetup{
    colorlinks=true,
    urlcolor=blue
}

\title{Imagine-then-Plan: Agent Learning from Adaptive \\ Lookahead with World Models}

\author{ 
    Youwei Liu$^{2,1}$\thanks{~This work was conducted while Youwei Liu was a remote research assistant at the Hong Kong Polytechnic University.}, ~Jian Wang$^{1,3 \dagger}$, ~Hanlin Wang$^{1}$, ~Beichen Guo$^{1}$, ~Wenjie Li$^{1}$ \\
    $^1$ Department of Computing, The Hong Kong Polytechnic University \\
    $^2$ Central South University  \\
    $^3$ College of Computer Science, Sichuan University \\
    \texttt{loyiv5477@gmail.com  ~ wangjian51@scu.edu.cn}
    \\ \texttt{\{hanlin-henry.wang,beichen.guo\}@connect.polyu.hk }
    \\
    \texttt{~~ cswjli@comp.polyu.edu.hk}
}

\begin{document}
\maketitle

\renewcommand{\thefootnote}{$\dagger$}
\footnotetext[1]{~Corresponding author. This work was mainly conducted at PolyU, while the author is now at Sichuan University.}
\setcounter{footnote}{0}
\renewcommand{\thefootnote}{\arabic{footnote}}

\begin{abstract}
Recent advances in world models have shown promise for modeling future dynamics of environmental states, enabling agents to reason and act without accessing real environments. 
Current methods mainly perform single-step or fixed-horizon rollouts, leaving their potential for complex task planning under-exploited.
We propose Imagine-then-Plan (\textbf{\model}), a unified framework for agent learning via lookahead imagination, where an agent's policy model interacts with the learned world model, yielding multi-step ``imagined'' trajectories. 
Since the imagination horizon may vary by tasks and stages, we introduce a novel adaptive lookahead mechanism by trading off the ultimate goal and task progress. 
The resulting imagined trajectories provide rich signals about future consequences, such as achieved progress and potential conflicts, which are fused with current observations, formulating a partially \textit{observable} and \textit{imaginable} Markov decision process to guide policy learning. 
We instantiate \texttt{ITP} with both training-free and reinforcement-trained variants. 
Extensive experiments across representative agent benchmarks demonstrate that \texttt{ITP} significantly outperforms competitive baselines. 
Further analyses validate that our adaptive lookahead largely enhances agents' reasoning capability, providing valuable insights into addressing broader, complex tasks. Our code and data are publicly available at: \noindent\faGithub\ \url{https://github.com/loyiv/ITP}.

\end{abstract}

\section{Introduction}

The emergence of Large Language Models (LLMs) has sparked a paradigm shift in autonomous agents, enabling them to reason and interact across a wide range of digital and physical environments~\citep{li2023camel,wang2024voyager,fung2025embodied}.
LLM-based agents primarily leverage immediate observations and historical interaction traces to facilitate decision-making~\citep{yao2023react,shinn2023reflexion}. 
Despite the impressive performance achieved, most agents remain constrained by \textit{shallow grounding}, a state where they perceive the environment but lack a deep, causal understanding of how their current actions will ultimately reshape the environment. 
Without the ability to project into the future, agents are prone to catastrophic failures, discovering erroneous actions or state conflicts only after they have been irreversibly executed.

\begin{figure}[t!]
    \centering
    \includegraphics[width=0.99\linewidth]{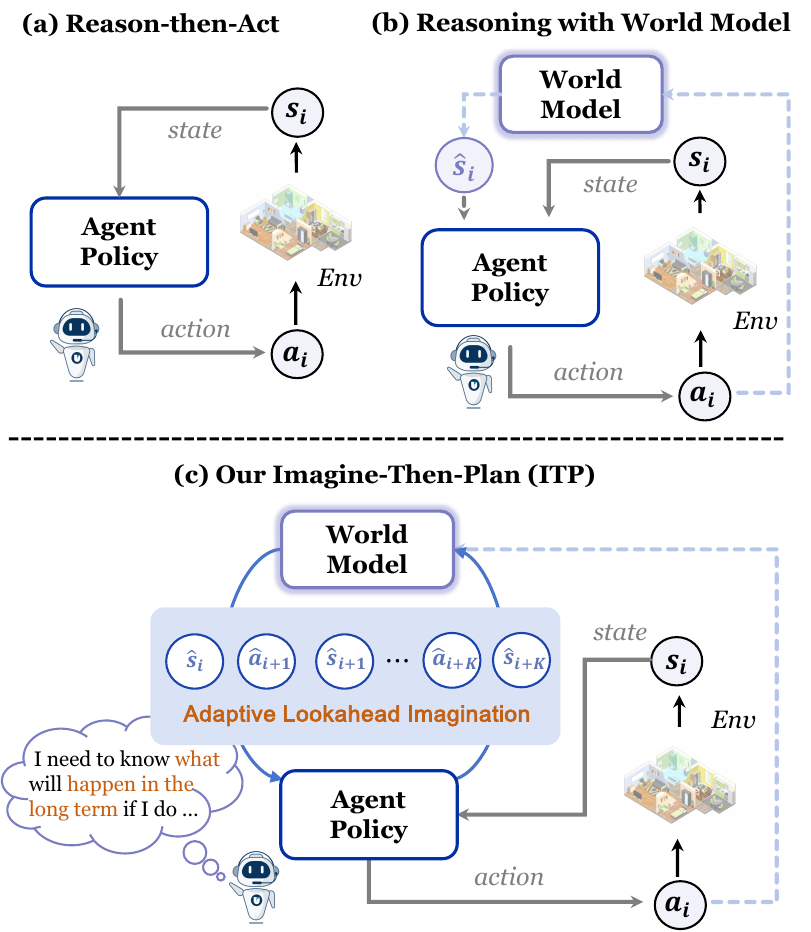}
    \caption{Comparison between our \textbf{\model} framework and conventional agent learning frameworks.}
    \label{fig:intro_idea}
    \vspace{-9pt}
\end{figure}

To bridge this grounding gap, world models~\citep{lecun2022path,hafner2023mastering} that focus on future state modeling have emerged as a promising solution, enabling agents to simulate environmental dynamics and ``rehearse'' actions in a mental sandbox~\citep{wang-etal-2025-world}. 
We focus on textual world models that predict decision-relevant future observations or state consequences, rather than continuous physical dynamics.
Contemporary methods typically employ world models for single-step verification or with fixed-horizon rollouts~\cite{hao-etal-2023-reasoning,qiao2024agent,zhang2025agent}. 
Such rigid strategies are often suboptimal, as they fail to capture long-term dependencies in complex tasks (e.g., household embodied tasks).
Furthermore, they are prone to suffer from high computational costs with unnecessary deep rollouts for trivial actions. 
A truly intelligent agent should be deliberative, allocating deep foresight adaptively to resolve potential state conflicts and account for long-term dependencies in pivotal, high-stakes decisions.

In this paper, we propose \texttt{\textbf{Imagine-then-Plan}} (\textbf{\model}), a framework that empowers agents to perform task planning with world model-based foresight. 
The core of \model is to move beyond passive observation and perform a proactive, deliberative ``rehearsal'' phase, where decisions are conditioned on both present observations and potential futures.
This requires the agent to internally imagine multi-step future trajectories by looking ahead. 
Unlike previous rigid approaches, \model introduces an \textbf{adaptive lookahead} mechanism that dynamically scales the imagination horizon by trading off the ultimate goal and estimated task progress. 
More importantly, this shift necessitates a new conceptualization of the agent's decision-making process. We move beyond the Partially Observable Markov Decision Process (POMDP)~\citep{aastrom1965optimal,song-etal-2024-trial,wang-etal-2025-steca} toward a Partially Observable and Imaginable MDP (POIMDP). 
As illustrated in Figure~\ref{fig:intro_idea}, the agent's action policy is optimized over a dual-stream representation:
the concrete present (\textit{observable}) and the foresighted future (\textit{imaginable}). 
These imagined trajectories provide rich signals, such as anticipating goal progress or detecting potential bottlenecks, allowing the agent to close the loop between planning action sequences and estimating their possible consequences. 
This implicit feedback enables the agent to perform self-correction when necessary, significantly reducing the need for expensive interactions with the real environment.

We instantiate \model in two variants: a \textbf{training-free} (\textbf{$\texttt{ITP}_{\text{I}}$}) variant that uses reflection as the adaptive lookahead for plug-and-play enhancement of LLM agents, and a \textbf{reinforcement-trained} (\textbf{$\texttt{ITP}_{\text{R}}$}) variant that leverages imagined futures to optimize the agent policy more effectively and more efficiently.
Extensive experiments demonstrate that both variants significantly improve task success rates. 
Further analyses validate the vital role of our adaptive lookahead mechanism.

Our contributions are summarized as follows:
1) We conceptualize the partially observable and imaginable Markov decision process (POIMDP), laying a solid foundation for integrating imagined futures and historical interactions into agent decision-making.
2) We propose Imagine-then-Plan (\model), a framework that incorporates world model-based imagination with an adaptive lookahead mechanism, which provides deliberative guidance for action policy planning.
3) We demonstrate through training-free and reinforcement-trained variants that \model significantly improves the success rates of LLM-based agents, providing valuable insights into addressing complex, long-horizon tasks.

\section{Preliminaries}
\label{sec:prelim}

\paragraph{Problem Formulation.}



The reasoning process of LLM agents is often formulated as a Partially Observable Markov Decision Process (POMDP), defined by $(\mathcal{S}, \mathcal{A}, \mathcal{O}, \mathcal{T}, \mathcal{R})$. 
Here, $\mathcal{S}$ denotes the environment state space, $\mathcal{A}$ the action space, and $\mathcal{O}$ the observation space. $\mathcal{T}: \mathcal{S} \times \mathcal{A} \rightarrow \mathcal{S}$ represents the state transition function, and $\mathcal{R}: \mathcal{S} \times \mathcal{A} \rightarrow [0, 1]$ denotes the reward function that evaluates task performance. 
At each time step $t$, the agent receives an observation $o_t \in \mathcal{O}$ and selects an action $a_t \sim \pi_\theta(\cdot | h_t)$,
where $h_t = (o_1, a_1, \dots, o_t)$ denotes the interaction history.
Executing $a_t$ induces a transition to a new latent state $s_{t+1} \sim \mathcal{T}(s_t, a_t)$, from which the environment emits the next observation $o_{t+1}$.
The interaction terminates when the agent reaches a terminal state or exceeds a predefined maximum number of steps.

\paragraph{LLMs as Textual World Models.}



\begin{figure*}[t!]
    \centering
    \includegraphics[width=0.98\textwidth]{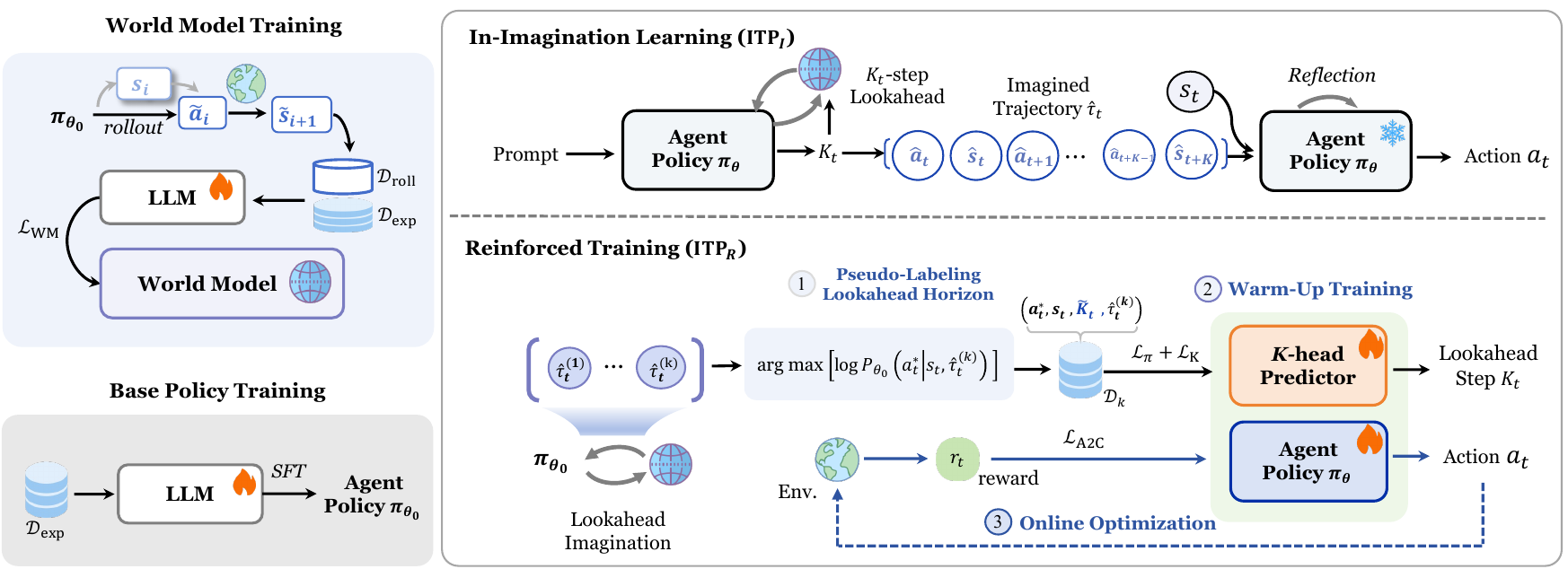}
    \caption{Overview of the proposed Imagine-then-Plan ({\textbf{\model}}) framework. It consists of two variants: (a) \textbf{$\texttt{ITP}_{\text{I}}$}, which is training-free and enables LLM agents to learn from the imagination at inference time. (b) \textbf{$\texttt{ITP}_{\text{R}}$}, which leverages an imagined future to optimize the action policy more effectively and more efficiently.
    }
    \label{fig:overview}
    \vspace{-8pt}
\end{figure*}

A world model is a predictive model of environment dynamics that estimates future states conditioned on actions~\cite{lecun2022path}.
In text-based environments, the environment state is typically represented as text and observable to the agent.
We treat textual observations as state representations, denoted as $s_t$ at time step $t$.
Under this formulation, LLMs can be interpreted as world models, as they capture transition regularities by predicting the next state given the current interaction context~\cite{zhang2025agent,li2025word}.
Concretely, a textual world model parameterized by an autoregressive LLM defines the conditional distribution $p_\phi(s_{t+1}| s_t, a_t)$.
The distribution is factorized at the token level, and the model generates the next state sequentially.
Such world models serve as powerful proxies for environment dynamics and enable planning in language-based agents.

\section{Method}
\label{sec:method}

We propose \texttt{\textbf{I}magine-\textbf{t}hen-\textbf{P}lan} (\textbf{\model}), a framework that equips LLM-based agents with adaptive lookahead via learned world models. 
As illustrated in Figure~\ref{fig:overview}, \textbf{\model} enables agents to condition decisions on both the observable present and imagined future trajectories.

\subsection{World Model Training}
As shown in Figure~\ref{fig:overview}, we first fine-tune a base LLM on expert demonstrations $\mathcal{D}_{\mathrm{exp}}$ to obtain an initial agent policy $\pi_{\theta_0}$.
This warm-up establishes basic capability to produce executable actions, serving as the foundation for the agent's exploration.
We ask the agent to perform rollouts in the environment, obtaining the rollout trajectories $\mathcal{D}_{\mathrm{roll}}$.
As introduced in \S~\ref{sec:prelim}, we learn a LLM-based world model $\mathcal{M}_{\phi}$ that approximates the environment dynamics $p_{\phi}(s{'} | s,a)$, 
where $s$ and $a$ denote the current state and action, respectively. $s{'}$ represents the next-step state.
To ensure the world model is grounded and robust to out-of-distribution actions, we train it on a joint dataset $\mathcal{D_{\mathrm{WM}}} = \mathcal{D}_{\mathrm{exp}} \cup \mathcal{D}_{\mathrm{roll}}$.

The world model $\mathcal{M}_{\phi}$ is optimized by minimizing the negative log-likelihood as follows:
\begin{equation}
    \mathcal{L}_{\text{WM}}(\phi) = - \mathbb{E}_{(s{'}, s, a)\sim\mathcal{D_{\mathrm{WM}}}} \bigl[\log p_{\phi}(s{'} | s,a)\bigr].
\end{equation}
We provide detailed training setup, dataset scale, and computational costs in Appendix~\ref{app:wm_training_details}.

\subsection{Lookahead Imagination and POIMDP}

To integrate world-model foresight into decision making, we extend the standard POMDP (as introduced in \S\ref{sec:prelim}) to a \textbf{P}artially \textbf{O}bservable and \textbf{I}maginable \textbf{M}arkov \textbf{D}ecision \textbf{P}rocess (\textbf{POIMDP}).
Under this formulation, the agent is endowed with a lookahead imagination operator induced by the learned world model.
POIMDP does not replace the underlying POMDP or modify the environment dynamics. Instead, it augments the agent-side information state with an internally imagined trajectory before action execution. When the lookahead horizon is selected as zero, the imagined trajectory is empty, and the decision reduces to the standard POMDP policy.

Formally, given the current observed state $s_t$ and a lookahead horizon $K_t \in \{0, 1, \dots, K_{\max}\}$ at each time step $t$, the agent policy $\pi_\theta$ and the world model $\mathcal{M}_{\phi}$ interact for $K_t$ steps within a ``mental sandbox''. 
This imagination process is given by:
\begin{equation}
    \xrightarrow{\pi_\theta} \hat{a}_{t} \xrightarrow{\mathcal{M}_\phi} \hat{s}_{t+1} \xrightarrow{\pi_\theta} \dots \xrightarrow{\pi_\theta} \hat{a}_{t+K_t-1} \xrightarrow{\mathcal{M}_\phi} \hat{s}_{t+K_t}.
\end{equation}
This yields an imagined future trajectory
$\hat{\tau}^{(K_t)}_t=\{(\hat{a}_{t+i},\hat{s}_{t+i+1})\}_{i=0}^{K_t-1}$,
where
$\hat a_{t+i}\sim\pi_\theta(\cdot | s_t,\hat\tau_t^{(i)})$
and
$\hat s_{t+i+1}\sim\mathcal{M}_{\phi}(\cdot | \hat s_{t+i},\hat a_{t+i})$.
The objective of the agent policy is to yield an appropriate action $a_t$ conditioned on both the observable state $s_t$ and the imagined future $\hat{\tau}^{(K_t)}_t$.
As such, our POIMDP formulates the policy decision as follows:
\begin{equation}
    a_t \sim \pi_\theta(\cdot \mid s_t, \hat{\tau}^{(K_t)}_t). 
\end{equation}
This allows the agent to anticipate goal progress or detect potential bottlenecks before generating the next action, enabling the agent to perform self-correction when necessary.

\subsection{Planning with Adaptive Lookahead}
For effective task planning, a key challenge is determining how far the agent should imagine.
Short-horizon imagination may miss long-term dependencies, while excessive rollouts can amplify model errors and incur unnecessary computation.
To resolve this, we aim to adaptively select the imagination horizon $K_t$ based on the estimated task progress against the ultimate goal.
We instantiate \textbf{\model} via two distinct variants to provide both flexibility and optimization: 
(\romannumeral1) \textbf{$\texttt{ITP}_{\text{I}}$}, which is an inference-time method that learns from the imagination, and (\romannumeral2) \textbf{$\texttt{ITP}_{\text{R}}$}, which is a reinforcement-trained method that jointly optimizes the lookahead horizon selection and action policy.

\subsubsection{In-Imagination Learning (\textbf{$\texttt{ITP}_{\text{I}}$})}

$\texttt{ITP}_{\text{I}}$ is a training-free variant that improves LLM agents at inference time. 
Both the policy and world model remain frozen, and the agent performs deliberative reasoning based on the current state $s_t$. 
At each step $t$, it follows a three-stage ``Imagine-then-Plan'' procedure:
1) \textit{Adaptive horizon selection}: The agent analyzes the instruction and state $s_t$ to choose an imagination horizon $K_t \in \{0, 1, \dots, K_{\max}\}$, assigning deeper foresight to critical decisions while avoiding unnecessary computation on trivial ones.
2) \textit{World-model imagination}: The agent interacts with $\mathcal{M}_{\phi}$ for $K_t$ steps to obtain a future trajectory $\hat{\tau}_t^{(K_t)}$.
3) \textit{Reflective policy generation}: Instead of directly executing the first imagined action, the agent treats $\hat{\tau}_t^{(K_t)}$ as implicit feedback for self-refinement.

Specifically, the agent reflects on the imagined trajectory to assess progress toward the goal and identify potential conflicts, bottlenecks, or catastrophic failures before acting in the real environment. 
It then refines its reflection and selects the next optimal action $a_t$ as follows:
\begin{equation}
    a_t \sim \pi_\theta(\cdot \mid s_t, \text{Reflect}(\hat{\tau}_t^{(K_t)})).
\end{equation}
By grounding decisions in the imagined future, $\texttt{ITP}_{\text{I}}$ turns passive observation into proactive, deliberative learning, improving task success without additional training.

\subsubsection{Reinforced Training (\textbf{$\texttt{ITP}_{\text{R}}$})}

$\texttt{ITP}_{\text{R}}$ aims to explicitly learn when and how long to imagine.
We augment the agent with a lightweight $K$-head predictor $P_\theta(K_t | s_t)$, which is a linear layer built on top of a backbone LLM that predicts distributions over imagination horizons. The action policy and predictor are optimized jointly with the following three stages.

\paragraph{Stage 1: Pseudo-Labeling Lookahead Horizon.}
A key obstacle for learning adaptive lookahead is that expert trajectories provide only $(s_t, a_t^*)$ pairs but do not specify the ``right'' lookahead horizon.
We therefore construct pseudo labels using the world model $\mathcal{M}_{\phi}$ and the initial agent policy $\pi_{\theta_0}$.
Specifically, we use teacher-forced expert actions to rollout on the frozen world model $\mathcal{M}_{\phi}$ by looking ahead one step and obtain future states, from which we derive lookahead-conditioned trajectories $\{\hat \tau_t^{(k)}\}_{k=0}^{K_{\max}}$. 
We then score each candidate step by the log-likelihood of expert actions under $\pi_{\theta_0}$, and select the optimal lookahead step by:
\begin{equation}
\tilde{K}_t=\arg\max_{0\le k\le K_{\max}}
\Big[\log p_{\theta_0}(a_t^* \mid s_t,\hat \tau_t^{(k)})-\lambda_K\,k\Big],
\label{eq:select_k}
\end{equation}
where $\lambda_K$ is a hyperparameter controlling the lookahead penalty. 
Based on the selection criteria in Eq. (\ref{eq:select_k}), we obtain a dataset $\mathcal{D}_K$ containing the pseudo-labels of the optimal lookahead step for each action in the expert trajectory.

\paragraph{Stage 2: Warm-Up Training.}
Starting from the initial agent policy $\pi_{\theta_0}$, we further fine-tune the agent to jointly (i) act under lookahead-conditioned pseudo-trajectories and (ii) predict the required lookahead step.
Specifically, given labeled tuples $\{(s_t, a_t^*, \tilde K_t)\}$, we condition the agent policy on $\hat \tau_t^{(\tilde K_t)}$ and train
$\pi_{\theta}(a_t | s_t, \hat \tau_t^{(\tilde K_t)})$ to imitate the expert action $a_t^*$, with a standard negative log-likelihood loss $\mathcal{L}_{\pi}(\theta)$. 
Meanwhile, we train the $K$-head predictor to estimate the pseudo label $\tilde K_t$, with a similar negative log-likelihood loss $\mathcal{L}_{K}(\theta)$. 
Our warm-up training is given by:
\begin{equation}
\mathcal{L}_{\mathrm{WT}}(\theta)=\mathcal{L}_{\pi}(\theta)+\eta\,\mathcal{L}_{K}(\theta).
\label{eq:warmup}
\end{equation}
where $\eta$ is a weighted coefficient. 
This yields (i) a competent agent policy that can generate reliable actions and (ii) an adaptive lookahead horizon predictor that approximates lookahead steps, providing a stable initialization for the subsequent online reinforcement optimization.

\paragraph{Stage 3: Online Optimization.}

To balance task performance and imagination cost, we further refine the agent policy through online reinforcement learning.
At each step $t$, the agent samples a lookahead step $K_t$ from the $K$-head predictor, invokes the frozen world model $\mathcal{M}_{\phi}$ to generate a $K_t$-step imagined trajectory $\hat{\tau}_t^{(K_t)}$, and subsequently samples an action $a_t \sim \pi_\theta(\cdot \mid s_t, \hat{\tau}_t^{(K_t)})$.
As illustrated in Figure~\ref{fig:overview}, we employ a reward function that augments the environment reward $r_{\mathrm{env}}$ with explicit penalties for computational and interaction overhead, given by:
\begin{equation}
    r_{t+1}=r_{\mathrm{env}}-\lambda_K K_t-\lambda_{\mathrm{step}},
\label{eq:reward}
\end{equation}
where $\lambda_K$ is the lookahead penalty coefficient, and $\lambda_{\text{step}}$ is a factor discouraging reasoning redundancy.

Specifically, we utilize the Advantage Actor-Critic (A2C) algorithm~\cite{mnih2016asynchronous} to jointly optimize the action policy and the $K$-head parameters. 
The objective is decomposed into three components:
(i) an actor term
$\mathcal{L}_{\mathrm{act}}(\theta)\triangleq-\mathbb{E}\!\left[A_t\left(\log p_\theta(K_t\mid s_t)+\log \pi_\theta(a_t\mid s_t,\hat{\tau}_t^{(K_t)})\right)\right]$
that jointly updates the lookahead predictor and the agent policy; 
(ii) a Critic regression term
$\mathcal{L}_{\mathrm{value}}(\theta)\triangleq\mathbb{E}\!\left[(V_\theta(s_t)-\hat{V}_t)^2\right]$
that trains a Value-head to match the TD learning target; 
and (iii) an entropy regularizer
$\mathcal{L}_{\mathrm{ent}}(\theta)\triangleq-\;\mathbb{E}\!\left[\mathcal{H}(p_\theta(K_t \mid s_t))\right]$
that encourages sufficient exploration over the lookahead steps to prevent premature convergence to sub-optimal horizons.
The final training objective is:
\begin{equation}
\mathcal{L}_{\mathrm{A2C}}(\theta)
=
\mathcal{L}_{\mathrm{act}}(\theta)
+\alpha \mathcal{L}_{\mathrm{value}}(\theta)
+\beta \mathcal{L}_{\mathrm{ent}}(\theta),
\label{eq:a2c_loss}
\end{equation}
where $\alpha$ and $\beta$ are hyperparameters balancing value estimation and exploration.
Algorithm~\ref{alg:itpr_algorithm_appendix} shows the pseudocode of ITPR's training process.

At inference time, $\texttt{ITP}_{\text{R}}$ utilizes the learned $K$-head to perform adaptive lookahead, following a similar deliberative procedure as $\texttt{ITP}_{\text{I}}$. 
We provide a stage-wise analysis on the computational cost of $\texttt{ITP}_{\text{R}}$ in Appendix~\ref{app:itr_training_cost}.

\section{Experiments}
\label{sec:experiment}

\subsection{Experimental Settings}

\paragraph{Benchmarks.}
We evaluate \textbf{\model} on four representative agent benchmarks: \textbf{ALFWorld}~\citep{shridhar2020alfred} for embodied household tasks, \textbf{ScienceWorld}~\citep{Wang2022ScienceWorld} for simulated science experiments, \textbf{WebShop}~\citep{yao2022webshop} for long-horizon web navigation, and \textbf{StableToolBench}~\citep{guo-etal-2024-stabletoolbench} for multi-turn tool use.
Appendix~\ref{app:data} provides details of these benchmarks.

\paragraph{Backbone Models.}
For a fair comparison, all methods use the same backbone suite: Qwen2.5-7B~\citep{qwen2.5}, Qwen3-8B~\citep{qwen3}, and Llama-3.1-8B-Instruct~\cite{dubey2024llama}.
Unless otherwise stated, we use Qwen3-8B to train the world model, and evaluate other world-model backbones in \S~\ref{sec:impact_wm}.
All models are prompted using their official chat templates.

\begin{table*}[t!]
\centering
\resizebox{0.98\textwidth}{!}{
\begin{tabular}{l l ccccccc c c c}
\toprule
\multirow{2}{*}{\textbf{Type}} &
\multirow{2}{*}{\textbf{Method}} &
\multicolumn{7}{c}{\textbf{ALFWorld}} &
\multicolumn{2}{c}{\textbf{ScienceWorld}} &
\multicolumn{1}{c}{\textbf{WebShop}} \\
\cmidrule(l){3-9} \cmidrule(l){10-11} \cmidrule(l){12-12}
& & PICK & CLEAN & HEAT & COOL & LOOK & PICK2 & Overall & Seen & Unseen & Total \\
\midrule

\multicolumn{12}{c}{\textit{Backbone: Qwen2.5-7B}~\qwenicon} \\
\midrule
\multirow{4}{*}{Prompting}
& CoT
& 17.14 & 18.52 & 18.75 & 16.00 & 15.38 & 0.00 & 14.29 & 3.09 & 4.63 & 5.10 \\
& ReAct
& 20.00 & 22.22 & \underline{18.75} & 20.00 & \underline{23.08} & 0.00 & 17.14 & 8.24 & 9.93 & \underline{15.28} \\
& RAP
& \underline{40.00} & \textbf{33.33} & 6.25 & \textbf{32.00} & 15.38 & \underline{20.83}
& \underline{27.86} & \underline{10.30} & \underline{16.55} & 11.28 \\
\cmidrule(l){2-9} \cmidrule(l){10-12}
& \gr \textbf{$\texttt{ITP}_{\text{I}}$ (Ours)}
& \gr\textbf{65.71} & \gr\underline{25.93} & \gr\textbf{25.00} & \gr\underline{24.00} & \gr\textbf{30.77} & \gr\textbf{25.00}
& \gr\textbf{35.71} & \gr\textbf{16.49} & \gr\textbf{17.88} & \gr \textbf{20.10} \\
\cmidrule(l){1-12}
\multirow{4}{*}{Training}
& SFT
& 85.71 & 66.67 & 56.25 & 68.00 & 38.46 & 66.67 & 67.86 & 55.67 & 49.00 & 51.60 \\
& WKM
& 77.14 & 77.78 & 75.00 & \underline{76.00} & \textbf{76.92} & 75.00 & 76.43 & 54.12 & 56.29 & \underline{58.80} \\
& IWM
& \underline{90.60} & \underline{85.20} & \textbf{88.20} & \textbf{84.20} & 42.90 & \underline{76.90}
& \underline{82.80} & \underline{60.82} & \underline{57.61} & 56.20 \\
\cmidrule(l){2-9} \cmidrule(l){10-12}
& \bl \textbf{$\texttt{ITP}_{\text{R}}$ (Ours)}
& \bl\textbf{94.29} & \bl\textbf{88.89} & \bl\underline{87.50} & \bl 53.84 & \bl\underline{76.00} & \bl\textbf{91.67}
& \bl\textbf{85.07} & \bl\textbf{62.58} & \bl\textbf{58.94} & \bl\textbf{60.20} \\
\midrule

\multicolumn{12}{c}{\textit{Backbone: Qwen3-8B}~\qwenicon} \\
\midrule
\multirow{4}{*}{Prompting}
& CoT
& 14.29 & 14.81 & 12.50 & 12.00 & 15.38 & 12.50 & 13.57 & 2.44 & 1.99 & 6.32 \\
& ReAct
& 25.71 & 22.22 & 12.50 & 12.00 & 7.69 & \underline{25.00} & 19.29 & 9.79 & 8.61 & \underline{18.62} \\
& RAP
& \underline{42.86} & \textbf{37.04} & \textbf{37.50} & \underline{16.00} & \underline{15.38} & 4.17
& \underline{28.57} & \underline{15.46} & \textbf{27.14} & 12.40 \\
\cmidrule(l){2-9} \cmidrule(l){10-12}
& \gr\textbf{$\texttt{ITP}_{\text{I}}$ (Ours)}
& \gr\textbf{82.86} & \gr\underline{25.93} & \gr\underline{12.50} & \gr\textbf{16.00} & \gr\textbf{23.08} & \gr\textbf{54.17}
& \gr\textbf{41.43} & \gr\textbf{20.61} & \gr\underline{19.86} & \gr \textbf{25.25} \\
\cmidrule(l){1-12}
\multirow{4}{*}{Training}
& SFT
& 71.43 & 70.37 & 68.75 & 72.00 & \underline{69.23} & 70.83 & 70.71 & 56.70 & 49.67 & 52.30 \\
& WKM
& 80.00 & 77.78 & 81.25 & 80.00 & \textbf{76.92} & 79.17 & 79.29 & \underline{60.31} & 47.68 & \underline{61.15} \\
& IWM
& \underline{85.71} & \underline{85.19} & \underline{87.50} & \underline{84.00} & 46.15 & \textbf{87.50} & \underline{82.14} & 59.27 & \underline{54.30} & 57.30 \\
\cmidrule(l){2-9} \cmidrule(l){10-12}
& \bl\textbf{$\texttt{ITP}_{\text{R}}$ (Ours)}
& \bl\textbf{97.14} & \bl\textbf{88.88} & \bl\textbf{93.75} & \bl\textbf{88.00} & \bl\textbf{76.92} & \bl\underline{79.17}
& \bl\textbf{88.57} & \bl\textbf{61.85} & \bl\textbf{56.95} & \bl\textbf{68.10} \\
\midrule

\multicolumn{12}{c}{\textit{Backbone: Llama-3.1-8B-Instruct}~\llamaicon} \\
\midrule
\multirow{4}{*}{Prompting}
& CoT
& 17.14 & 14.81 & \underline{18.75} & 16.00 & \underline{15.38} & 8.33 & 15.00 & 3.09 & 3.31 & 8.20 \\
& ReAct
& 22.86 & 22.22 & 6.25 & 24.00 & \textbf{23.08} & 25.00 & 21.43 & 9.27 & 13.24 & \underline{19.32} \\
& RAP
& \underline{25.71} & \underline{25.93} & 6.25 & \textbf{32.00} & 7.69 & \underline{25.00}
& \underline{22.86} & \underline{11.34} & \underline{17.21} & 16.05 \\
\cmidrule(l){2-9} \cmidrule(l){10-12}
& \gr\textbf{$\texttt{ITP}_{\text{I}}$ (Ours)}
& \gr\textbf{57.14} & \gr\textbf{37.04} & \gr\textbf{31.25} & \gr\underline{28.00} & \gr\textbf{23.08} & \gr\textbf{33.33}
& \gr\textbf{37.86} & \gr\textbf{19.58} & \gr\textbf{19.20} & \gr \textbf{24.30} \\
\cmidrule(l){1-12}
\multirow{4}{*}{Training}
& SFT
& 85.71 & 85.20 & 82.40 & 89.50 & \underline{85.70} & 53.80 & 79.28 & 57.21 & 50.33 & 47.30 \\
& WKM
& 85.71 & 85.19 & 75.00 & 80.00 & 38.46 & 79.17 & 77.86 & \underline{61.34} & 54.96 & \underline{65.58} \\
& IWM
& \underline{87.50} & \underline{88.90} & \underline{82.40} & \textbf{94.70} & \textbf{85.90} & \underline{84.60}
& \underline{85.90} & 57.56 & \underline{56.29} & 58.60 \\
\cmidrule(l){2-9} \cmidrule(l){10-12}
& \bl\textbf{$\texttt{ITP}_{\text{R}}$ (Ours)}
& \bl\textbf{88.57} & \bl\textbf{92.59} & \bl\textbf{93.75} & \bl\underline{92.00} & \bl 46.15 & \bl\textbf{91.67}
& \bl\textbf{87.14} & \bl\textbf{63.91} & \bl\textbf{57.61} & \bl\textbf{67.50} \\
\bottomrule
\end{tabular}
}
\caption{Evaluation of task success rates (\%) across ALFWorld, ScienceWorld, and WebShop benchmarks. \textbf{Bold} and \underline{underlined} values denote the best and second-best performance within each backbone model group, respectively.}
\label{tab:main_results}
\vspace{-8pt}
\end{table*}

\paragraph{Baseline Methods.}

We compare \textbf{\model} against two categories of baselines. 
(1) \textit{Prompting-based} methods:
\textbf{CoT}~\citep{wei2022chainofthought} elicits reasoning capabilities by prompting the agent with step-by-step rationales.
\textbf{ReAct}~\citep{yao2023react} interleaves reasoning and action to solve interactive tasks.
\textbf{RAP}~\citep{hao-etal-2023-reasoning} leverages the LLM as both a world model and a policy model, employing Monte Carlo Tree Search to perform planning.
(2) \textit{Training-based} methods:
\textbf{SFT}~\citep{chen2023fireact} conducts behavioral cloning on expert trajectories.
\textbf{WKM}~\citep{qiao2024agent} trains a parametric world knowledge model that provides global task priors and local dynamic state knowledge to guide planning.
\textbf{IWM}~\cite{zhang2025agent} augments imitation learning with an implicit world-modeling objective to encourage the policy to internalize environment dynamics. All baselines are evaluated using the same benchmark splits, executable
action space, state--action serialization, backbone family, and evaluation
metric as \model. We additionally consider stronger search-based,
world-model-based, and trajectory-optimization methods. Their complete
results and reproduction details are provided in
Appendix~\ref{app:strong_baselines}.

\paragraph{Evaluation Metrics.}

We adopt success rate (\textbf{SR}) as the evaluation metric, defined as the percentage of episodes that successfully achieve the task goal~\citep{feng2025group,wang2025spa}. 
For ALFWorld, we report SR on each task, as well as the overall SR. 
For ScienceWorld, we report SR on both the \textit{seen} and \textit{unseen} test splits. 
Following~\citet{guo-etal-2024-stabletoolbench}, we report solvable pass rate (\textbf{SoPR}) and solvable win rate (\textbf{SoWR}) on the solvable subset for StableToolBench, with the detailed definitions provided in Appendix~\ref{app:data}.

\begin{table}[t]
\centering
\small
\setlength{\tabcolsep}{7pt}
\resizebox{0.98\linewidth}{!}{
\begin{tabular}{llcc}
\toprule
\textbf{Type} & \textbf{Method} & \textbf{SoPR (\%)} & \textbf{SoWR (\%)} \\
\midrule
\multirow{3}{*}{Prompting}
& ReAct & 26 & 22 \\
& RAP & \underline{28} & \underline{26} \\
& \gr \textbf{$\texttt{ITP}_{\text{I}}$ (Ours)} & \gr \textbf{35} (\textcolor{red!60!black}{$\scriptscriptstyle +7$}) & \gr \textbf{28} (\textcolor{red!60!black}{$\scriptscriptstyle +2$}) \\
\midrule
\multirow{3}{*}{Training}
& SFT & 42 & 36 \\
& IWM & \underline{44} & \underline{36} \\
& \bl \textbf{$\texttt{ITP}_{\text{R}}$ (Ours)} & \bl \textbf{68} (\textcolor{red!60!black}{$\scriptscriptstyle +24$}) & \bl \textbf{54} (\textcolor{red!60!black}{$\scriptscriptstyle +18$}) \\
\bottomrule
\end{tabular}
}
\caption{Evaluation of solvable pass rate (SoPR) and solvable win rate (SoWR) on StableToolBench using Qwen3-8B~\protect\qwenicon~as the backbone model.}
\label{tab:stabletoolbench_main}
\vspace{-8pt}
\end{table}

\begin{figure}[th]
    \centering
    \includegraphics[width=0.96\linewidth]{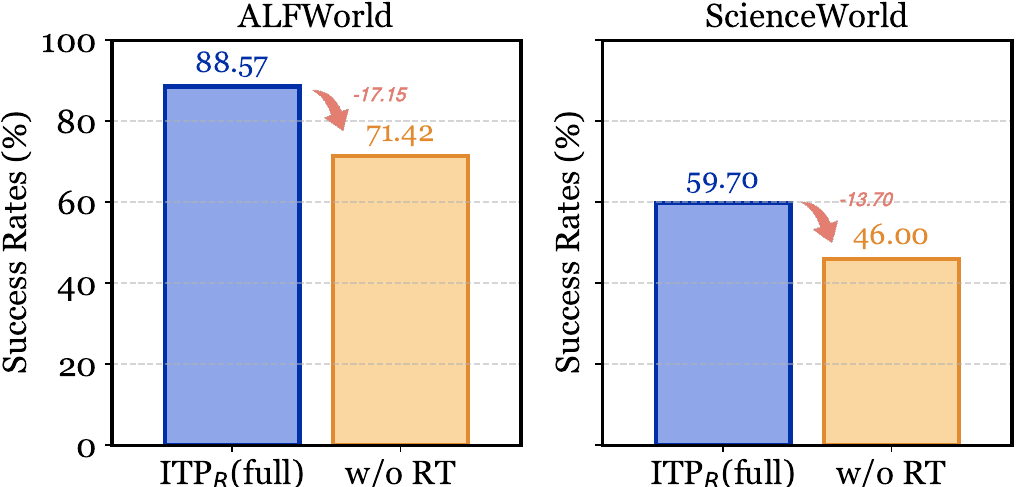}
    \caption{Ablation results of $\texttt{ITP}_{\text{R}}$ on ALFWorld and ScienceWorld benchmarks.}
    \label{fig:ablation}
    \vspace{-8pt}
\end{figure}

\paragraph{Implementation Details.}

To ensure a rigorous and fair comparison, we maintain a consistent training protocol across \model and all baseline methods. 
The maximum lookahead horizon is configured based on task completion steps per benchmark.
For more details, please refer to Appendix~\ref{appendix:implement}.

\subsection{Main Results}

\paragraph{How does \model perform across different benchmarks and backbone models?}
As shown in Table~\ref{tab:main_results}, without any additional training, \textbf{$\texttt{ITP}_{\text{I}}$} substantially enhances zero-shot performance compared to strong prompting baselines, such as ReAct and RAP.
The trained variant \textbf{$\texttt{ITP}_{\text{R}}$} consistently surpasses all training-based baselines, achieving the highest success rate in every backbone group (e.g., 88.57\% with Qwen3-8B on ALFWorld).
These results verify that \textbf{\model} achieves consistent gains across tested benchmarks and backbone models.

\begin{figure*}[t!]
  \centering
  \begin{subfigure}[t]{0.38\textwidth}
    \centering
    \includegraphics[width=\linewidth]{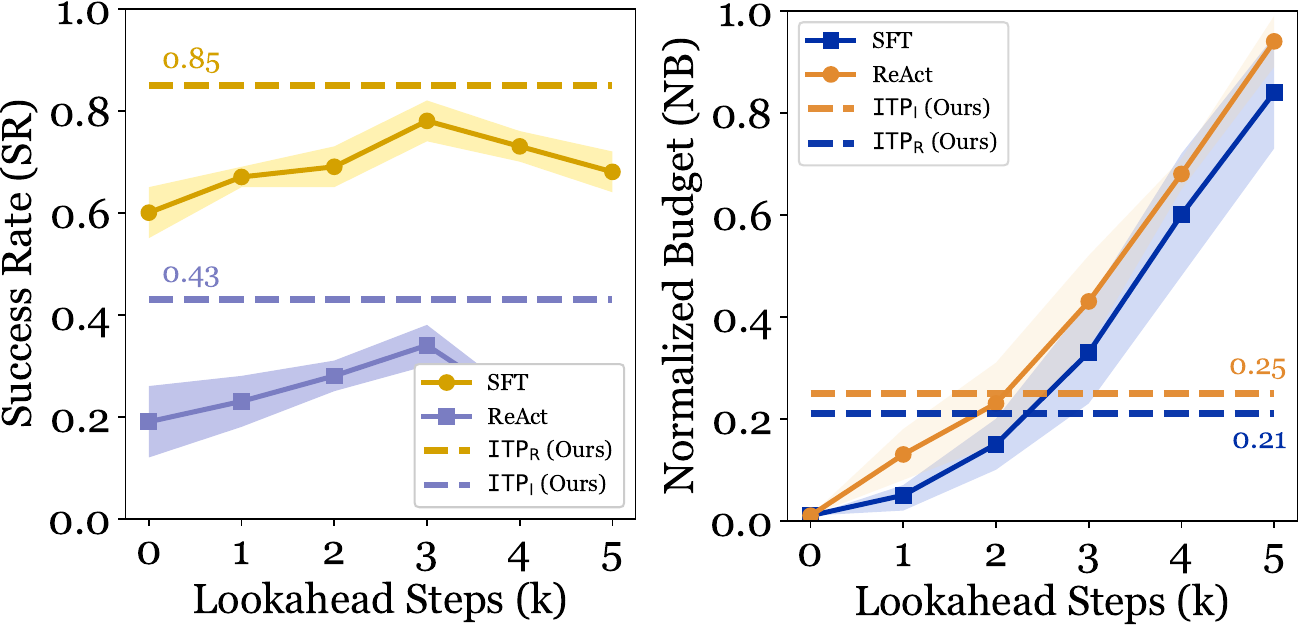}
    \caption{Adaptive lookahead vs. fixed-$k$ lookahead. \textit{Left}: success rate. \textit{Right}: normalized budget.}
    \label{fig:fixed_k}
  \end{subfigure}\hfill
  \begin{subfigure}[t]{0.29\textwidth}
    \centering
    \includegraphics[width=\linewidth]{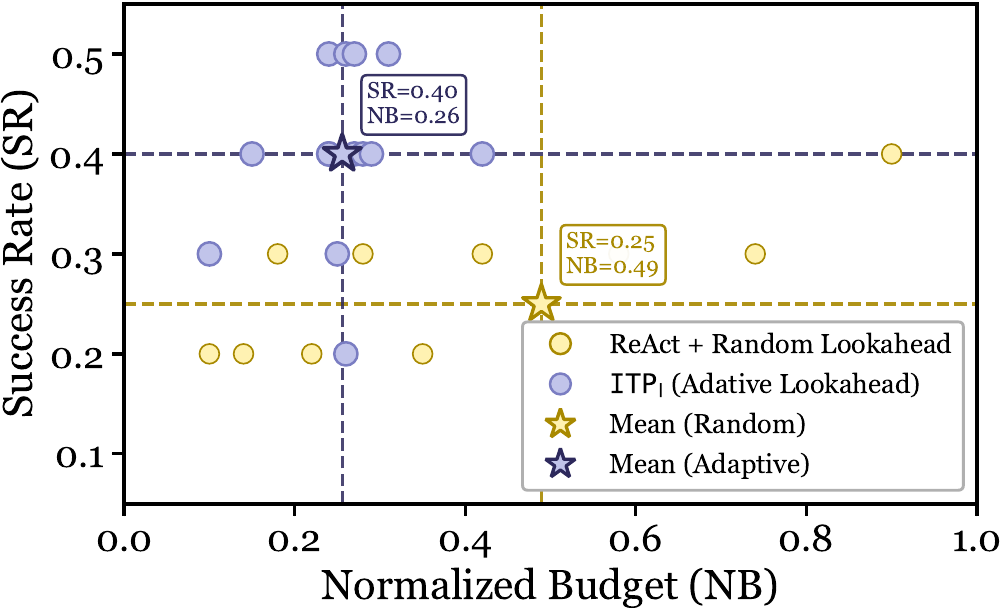}
    \caption{Our \textbf{$\texttt{ITP}_{\text{I}}$} vs. ReAct + Random Lookahead}
    \label{fig:itpi_sr_budget}
  \end{subfigure}\hfill
  \begin{subfigure}[t]{0.29\textwidth}
    \centering
    \includegraphics[width=\linewidth]{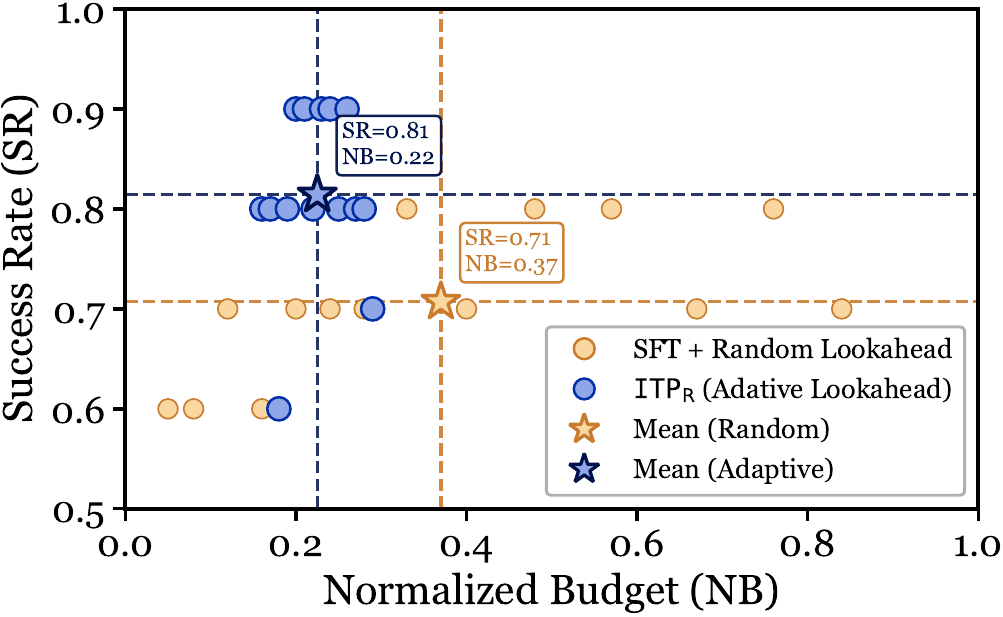}
    \caption{Our \textbf{$\texttt{ITP}_{\text{R}}$} vs. SFT + Random Lookahead}
    \label{fig:itpr_sr_budget}
  \end{subfigure}

  \vspace{-4pt}
  \caption{Performance-Cost comparison across lookahead strategies. (a): Our \textbf{adaptive} lookahead outperforms \textbf{fixed} lookahead with higher success rates and lower computational cost. (b) and (c): Both \textbf{$\texttt{ITP}_{\text{I}}$} and \textbf{$\texttt{ITP}_{\text{R}}$} with the adaptive lookahead surpass baselines (ReAct and SFT) with a \textbf{random} lookahead strategy.}
  \label{fig:lookahead_tradeoff}
  \vspace{-8pt}
\end{figure*}

\paragraph{Do \model’s performance gains stem from lookahead with world models?}
The effectiveness of our approach is twofold.
First, \textbf{$\texttt{ITP}_{\text{I}}$} uses the same backbone LLM as the prompting baselines, so its gains come solely from conditioning actions on world-model rollouts at inference time, isolating the value of explicit lookahead.
Second, \textbf{$\texttt{ITP}_{\text{R}}$} significantly outperforms training-based alternatives such as IWM and WKM.
Its strongest results (e.g., 63.91\% on ScienceWorld test-seen with Llama3.1-8B) indicate that combining policy learning with selective adaptive lookahead yields benefits beyond static or implicit methods.

\paragraph{Ablation Study.}

To assess the contribution of $\texttt{ITP}_{\text{R}}$, we remove its online reinforced training (\textbf{w/o RT}) for an ablation study.
In this setting, the model relies solely on initial training and does not learn when to invoke the world model.
We use Qwen-3-8B for both the agent policy and world model, and evaluate on ALFWorld and ScienceWorld.
As shown in Figure~\ref{fig:ablation}, removing RT substantially degrades performance: ALFWorld drops from 88.57\% to 71.42\%, and ScienceWorld from 59.70\% to 46.00\%.
This confirms that the online reinforced training is a core component rather than a minor training detail.
The gap further suggests that supervised learning provides only a basic capability, while reinforcement optimization is crucial for learning when to ``imagine'' and use the world model efficiently on complex tasks.

\subsection{Benefits of Adaptive Lookahead}

\paragraph{Superior performance-efficiency trade-off over Fixed Lookahead.}
We first compare \textbf{\model} with fixed-$k$ lookahead, where the agent always imagines a constant horizon $k$.
We measure computational cost by the total episode tokens
$T=\sum_t\bigl(T^{(t)}(\pi_{\theta})+T^{(t)}(\mathcal{M}_{\phi})\bigr)$,
and define a \textbf{Normalized Budget (NB)} metric as follows:
$$
\text{NB}(k)=\frac{\bar{T}(k)-\bar{T}(0)}{\bar{T}(K_{\max})-\bar{T}(0)},
$$
which rescales the average token cost $\bar{T}(k)$ to $[0,1]$. 
Appendix~\ref{appendix:implement} reports the setting of $K_{\max}$.

As shown in Figure~\ref{fig:fixed_k}, fixed-$k$ lookahead is brittle: success rate peaks at a moderate $k$ and then declines, while cost rises sharply with $k$.
In contrast, \textbf{\model}'s adaptive lookahead achieves higher success rates with a substantially lower budget, avoiding both the need for global horizon tuning and the high cost of large lookahead steps.

\paragraph{Higher success rates with lower cost than Random Lookahead.}
Beyond fixed-horizon lookahead, we compare \textbf{\model} with \textit{random lookahead}, which samples $K_t$ independently at each step.
This isolates the effect of adaptive horizon selection from simply using a varying horizon.
We evaluate the performance-cost trade-off on ALFWorld, using Qwen3-8B as both the policy and world model.
We run 140 tasks grouped into 14 folds and report fold-averaged SR and NB, where each point in Figure~\ref{fig:itpi_sr_budget} and Figure~\ref{fig:itpr_sr_budget} corresponds to one fold.

Our adaptive lookahead consistently outperforms the random strategy, achieving higher SR with lower and more stable budgets.
This shows that the gains come from state-conditioned allocation of lookahead, rather than horizon variability.

\subsection{Impact of World Models}
\label{sec:impact_wm}

\begin{figure}[t!]
  \centering
  \begin{subfigure}[t]{0.49\columnwidth}
    \centering
    \includegraphics[width=\linewidth]{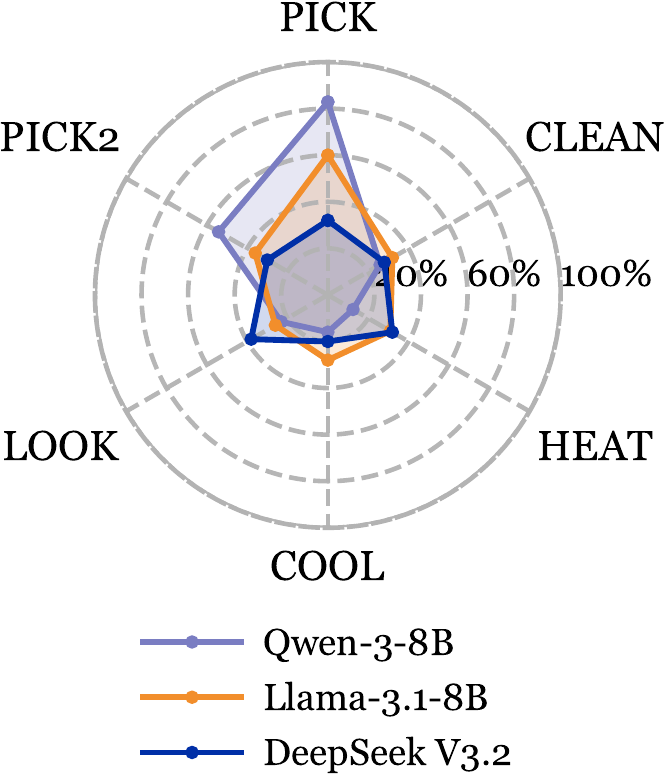}
    \caption{$\texttt{ITP}_{\text{I}}$ (Ours)}
    \label{fig:wm_itpi_radar}
  \end{subfigure}
  \hfill
  \begin{subfigure}[t]{0.49\columnwidth}
    \centering
    \includegraphics[width=\linewidth]{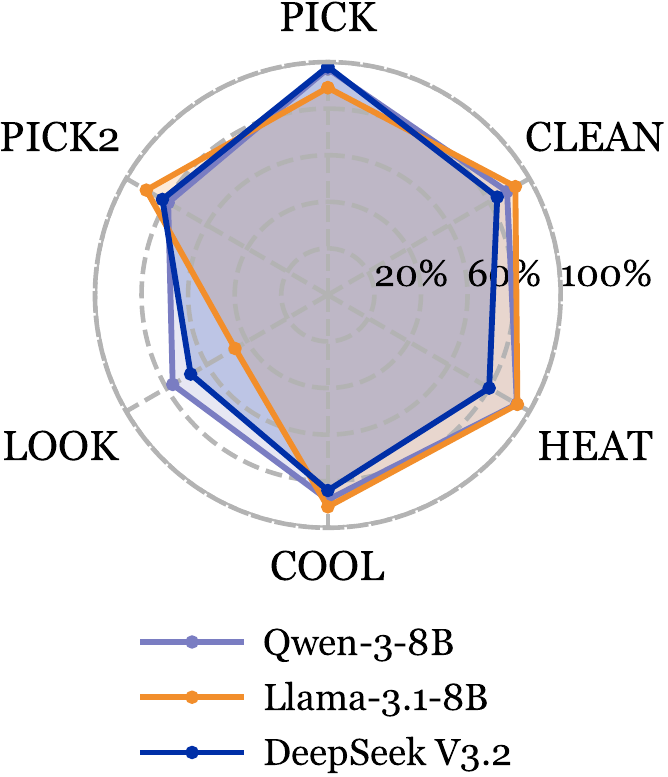}
    \caption{$\texttt{ITP}_{\text{R}}$ (Ours)}
    \label{fig:wm_itpr_radar}
  \end{subfigure}
  \caption{Impact of different world-model backbones. We report success rates of (a) $\texttt{ITP}_{\text{I}}$ and (b) $\texttt{ITP}_{\text{R}}$ across six task types on ALFWorld.}
  \label{fig:wm_impact_radar}
  \vspace{-8pt}
\end{figure}

\paragraph{Does the choice of world-model backbones affect ultimate performance?}
We evaluate the sensitivity of \textbf{\model} to the underlying world models by varying the world-model backbones among Qwen3-8B, Llama-3.1-8B, and DeepSeek-V3.2~\citep{deepseek2025v32}. We fix the agent policy to Qwen3-8B and evaluate on the ALFWorld benchmark.
As illustrated in Figure~\ref{fig:wm_impact_radar}, the choice of backbone is most consequential in the training-free ($\textbf{\model}_{\text{I}}$) setting. 
Specifically, despite its superior scale, the zero-shot DeepSeek-V3.2 exhibits a noticeable performance deficit, likely due to a lack of domain-specific alignment with ALFWorld's state-transition dynamics. 
In contrast, Qwen and Llama backbones maintain robust success rates, suggesting better inherent compatibility. 
Crucially, $\textbf{\model}_{\text{R}}$ substantially narrows this gap. After optimizing the use of the DeepSeek-V3.2 world model, \textbf{\model}'s performance becomes highly competitive. 
These results demonstrate that $\texttt{ITP}$ is model-agnostic and can effectively distill world-modeling capabilities from various architectures.
Furthermore, targeted world-modeling training is essential in specific interactive environments.



\begin{figure}[t!]
  \centering
  \begin{subfigure}[t]{0.50\columnwidth}
    \centering
    \includegraphics[width=\linewidth]{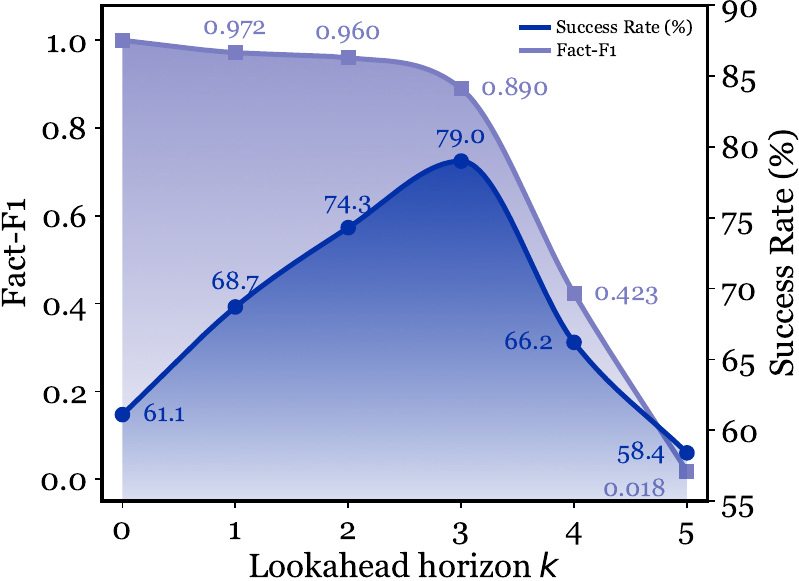}
    \caption{Fact-F1 of the world model vs. task success rate.}
    \label{fig:factf1_horizon}
  \end{subfigure}
  \hfill
  \begin{subfigure}[t]{0.48\columnwidth}
    \centering
    \includegraphics[width=\linewidth]{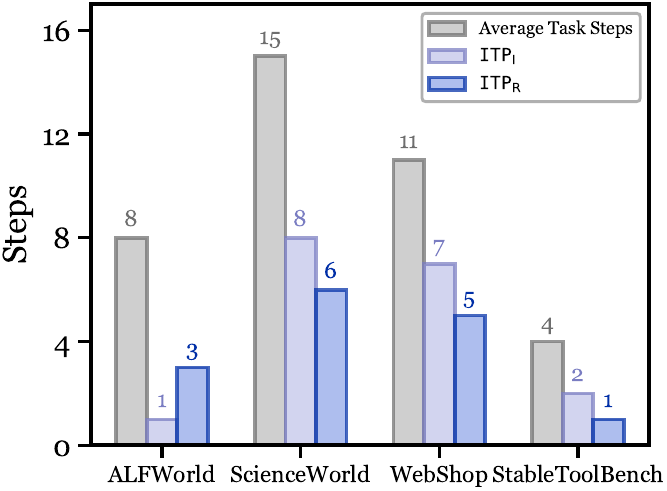}
    \caption{Statistics of average lookahead vs. golden steps.}
    \label{fig:k_dist_benchmarks}
  \end{subfigure}
  \caption{World model's reliability and lookahead horizon statistics on ALFWorld with Qwen3-8B~\protect\qwenicon. (a) Longer lookahead horizon reduces \textit{Fact-F1} of imagined states, while success rate peaks at $k{=}3$. (b) Longer-horizon tasks induce larger lookahead steps.}
  \label{fig:reliability_horizon_analysis}
  \vspace{-8pt}
\end{figure}

\paragraph{How reliable is the world model across lookahead horizons?}
We evaluate adaptive lookahead from two perspectives: world-model reliability and the selection of the lookahead horizon ($k$). We use \textit{Fact-F1} as the reliability metric, computed by comparing canonicalized atomic facts extracted from predicted versus ground-truth states.
As shown in Figure~\ref{fig:factf1_horizon}, \textit{Fact-F1} degrades as the horizon increases due to compounding errors. Interestingly, the agent's success rate follows a bell curve, peaking at $k{=}3$, which suggests a trade-off between foresight and prediction accuracy. 
Figure~\ref{fig:k_dist_benchmarks} further demonstrates that \textbf{\model}'s learned $k$ distributions adapt to task complexity: longer-horizon environments (e.g., ScienceWorld) favor larger lookahead values. 
These results validate \textbf{\model}'s core design: effective planning requires dynamically anchoring the lookahead horizon within the world model's reliable predictive range.


\section{Related Work}

\paragraph{LLM-based Agents.}
LLM-based agents use language models as policies that map instructions and partial observations to actions.
A significant line of work formulates agent learning as trajectory- or step-level optimization. 
For example, ETO~\citep{song-etal-2024-trial} frames learning as exploration-based trajectory optimization, while IPR~\citep{xiong-etal-2024-watch} and E$^2$CL~\citep{wang-etal-2024-e2cl} refine agent behavior via iterative revision and correction signals.
More recent post-training methods further improve robustness and generalization through reflective updates, such as Agent-R~\citep{yuan2025agent}, STeCa~\citep{wang-etal-2025-steca}, and AgentRefine~\citep{fu2025agentrefine}.
However, these methods primarily focus on learning from historical traces, often leaving the agent in a state of ``shallow grounding'' where it lacks active foresight to anticipate future environmental shifts before execution.

\paragraph{World Models for Planning.}
World models provide a ``mental sandbox'' for agents, enabling model-based decision-making by predicting environment dynamics~\citep{lecun2022path,xiang2024language,wang-etal-2025-world,liu2026comap}. 
Recent research has increasingly positioned LLMs as either implicit or explicit world models for task planning. 
For instance, RAP~\citep{hao-etal-2023-reasoning} treats LLM reasoning as a planning process over an implicit state space, while other works explicitly construct world models to simulate future states and verify plan feasibility~\citep{guan2023leveraging}. 
To improve simulation fidelity, methods like WKM~\citep{qiao2024agent}, DMWM~\citep{wang2025dmwm}, and IWM~\citep{zhang2025agent} incorporate world knowledge or historical experience to guide rollouts, while D2PO~\citep{wang-etal-2025-world} and internalizing strategies~\citep{chen2025internalizing} align planning behavior with environmental feedback. 
Further specialized work, such as WebEvolver~\citep{fang-etal-2025-webevolver}, explores world model-based imagination in web agents.
Unlike existing methods constrained by rigid, fixed-horizon planning, our work introduces an adaptive lookahead mechanism that dynamically adjusts the imagination horizon, enabling the agent to optimally navigate the trade-off between foresight reliability and computational overhead.

\section{Conclusion}

In this paper, we propose Imagine-then-Plan (\model), an agent learning framework that equips LLM-based agents with adaptive lookahead with world models. 
By extending the conventional Partially Observable MDP to a Partially Observable and Imaginable MDP, \model enables agents to explicitly reason over both the present and foresighted trajectories, addressing the shallow grounding limitation of reactive decision-making. 
We instantiate \model into both an inference-time variant and a reinforcement-trained variant.
Extensive experiments demonstrate that our approach significantly improves task success and robustness across domains. 
Further analyses validate the critical role of the adaptive control of the imagination horizon. 
We believe that  \model provides a principled advancement toward more deliberative utilization of world models for autonomous agent learning.

\section*{Limitations}
While our approach demonstrates superior performance compared to baseline methods, it is important to acknowledge the limitations as follows: 
(1) Current evaluation primarily focuses on interactive text-based benchmarks. While these environments provide a rigorous test of long-horizon reasoning, they do not fully capture the challenges of multimodal or real-world physical environments. 
The transition from textual state descriptions to visual or sensorimotor observations may introduce additional noise that could affect the stability of the adaptive lookahead mechanism.
(2) Although our adaptive lookahead mechanism is designed to optimize efficiency by scaling the imagination horizon, the use of world models inherently introduces higher inference-time overhead compared to purely reactive agents. 
While higher success rates in high-stakes tasks often justify this trade-off, further optimization, e.g., via speculative decoding~\citep{xia-etal-2023-speculative} or balancing exploration and exploitation~\citep{wei2012coaching}, is needed for real-time applications.
We will leave these directions as our future work.

\section*{Ethics Statement}
We strictly follow the protocols governing the academic use of all LLMs. Our study is conducted in simulated, text-based benchmarks and involves no human subjects or personally identifying information. We cite and comply with the licenses of all models, datasets, and software used. 
We acknowledge that world-model-based lookahead may introduce potential risks if transferred beyond our simulated evaluation settings: prediction errors and hallucinated rollouts could lead to unsafe or unintended actions in open-world tool-use systems or robots, and the generation of imagined trajectories increases inference-time computation and energy cost. 
Additionally, while AI assistants (e.g., Cursor and ChatGPT) were partially utilized for coding and linguistic refinement, we affirm that all core content and findings in this paper are the original work of the authors.

\section*{Acknowledgements}
This work was supported by the Fundamental Research Funds for the Central Universities (YJ202620 and 2026SCUQJTX022B), the General Research Fund (GRF) of the Research Grants Council of Hong Kong (PolyU 15205325), and was supported in part by the PolyU Postdoc Matching Fund (4-W40Z).
The authors would like to thank the anonymous reviewers for their valuable feedback and constructive suggestions.

\bibliography{custom}

\clearpage
\FloatBarrier

\appendix
\section{Datasets and Preprocessing}
\label{app:data}

\subsection{Datasets}
We evaluate our method on four representative agent benchmarks:
ALFWorld\footnote{\url{https://github.com/alfworld/alfworld}}~\citep{shridhar2020alfred},
ScienceWorld\footnote{\url{https://github.com/allenai/ScienceWorld}}~\citep{Wang2022ScienceWorld},
WebShop\footnote{\url{https://github.com/princeton-nlp/WebShop}}~\citep{yao2022webshop},
and StableToolBench\footnote{\url{https://github.com/THUDM/StableToolBench}}~\citep{guo-etal-2024-stabletoolbench}.
Table~\ref{tab:benchmark_stats} presents task descriptions and data statistics. 

\paragraph{ALFWorld:}
ALFWorld is a text-based household embodied benchmark where an agent interacts with a simulated environment via natural-language observations and admissible text actions. Each episode specifies a goal instruction that can be instantiated from compositional templates, and success requires planning over multi-step action sequences under partial observability. 

\begin{itemize}
  \item \textsc{Pick:} to find an object of the desired type, pick it up, navigate to the correct location/receptacle, and place it there.
  \item \textsc{Clean:} to find the target object, pick it up, go to a sink/basin, wash it by turning on the faucet, then navigate to the target receptacle and place it.
  \item \textsc{Heat:} to find the target object, pick it up, go to a microwave, heat it by turning on the microwave, then place it at the specified location.
  \item \textsc{Cool:} to find the target object, pick it up, go to a fridge, cool it by placing it inside the fridge, then return and place it at the specified location.
  \item \textsc{Look:} to find the target object, locate a light source, turn it on, and examine the object with the light while holding it.
  \item \textsc{Pick2:} to find the first target object and place it at the destination, then find a second object of the same type, return to the destination, and place it together with the first one.
\end{itemize}

\paragraph{ScienceWorld:}
ScienceWorld is a text-based interactive science environment that evaluates an agent’s ability to solve procedural and reasoning-intensive tasks grounded in everyday scientific phenomena. Compared to household tasks, ScienceWorld typically involves longer horizons and requires the agent to combine information gathering, tool use, and multi-step experimentation.

\paragraph{WebShop:}
WebShop is a text-based web shopping benchmark in which an agent interacts with simulated e-commerce webpages to search for products that satisfy a user instruction. Compared with embodied environments, WebShop requires the agent to reason over longer and noisier textual observations such as search results, product titles, attributes, and descriptions, and to perform multi-step browsing, comparison, and selection before making a final purchase decision.

\paragraph{StableToolBench:}
StableToolBench is a multi-turn tool-use benchmark that evaluates an agent's ability to solve tasks by invoking executable tools under realistic interaction constraints. In contrast to text navigation environments, StableToolBench emphasizes the correctness and robustness of tool calling, intermediate execution feedback handling, and solvability-aware evaluation. Although episodes are typically shorter, each decision is more sensitive because an incorrect tool invocation may immediately derail downstream progress.

We employ Qwen3-8B as the backbone model for different methods. Following existing studies, we adopt solvable pass rate (\textbf{SoPR}) and solvable win rate (\textbf{SoWR}) as the evaluation metrics. 
SoPR measures the average task success on the ``solvable'' subset (i.e., queries that are answerable given the available tools/KB, where each instance is judged as solved/unsolved/uncertain and mapped to 1/0/0.5. 
SoWR is a pairwise metric on the same solvable subset, reporting how often our method wins against a fixed baseline. In particular, SoWR is computed on the solvable subset as the average head-to-head win rate against the baseline set (ReAct, RAP, SFT, and IWM), using the official evaluation prompts and tool API.

\begin{table}[t]
  \centering
  \footnotesize
  \setlength{\tabcolsep}{4pt}
  \renewcommand{\arraystretch}{1.12}
  \begin{tabularx}{\columnwidth}{@{}
    >{\raggedright\arraybackslash}p{1.85cm}
    >{\raggedright\arraybackslash}X
    >{\raggedright\arraybackslash}X
  @{}}
    \toprule
    \textbf{Domain} & \textbf{Task Description} & \textbf{Dataset Statistics} \\
    \midrule

    \multicolumn{3}{c}{\textbf{ALFWorld}} \\
    \midrule
    Household, text-based embodied tasks &
    Six compositional task families: \textsc{Pick}, \textsc{Clean}, \textsc{Heat}, \textsc{Cool}, \textsc{Look}, \textsc{Pick2}. &
    \begin{tabular}[t]{@{}l@{}}
      Training: 3,119 \\
      Test: 140 
    \end{tabular} \\
    \midrule

    \multicolumn{3}{c}{\textbf{ScienceWorld}} \\
    \midrule
    Elementary science curriculum in an interactive text environment &
    30 subtasks with many variations (entities, initial conditions, distractors, room layouts), partitioned following the benchmark protocol. &
    \begin{tabular}[t]{@{}l@{}}
      Training: 1,483 \\
      Test-Seen: 194 \\
      Test-Unseen: 151
    \end{tabular} \\

        \midrule

    \multicolumn{3}{c}{\textbf{WebShop}} \\
    \midrule
    Goal-driven web shopping in a text-based browsing environment &
    Product search, browsing, attribute comparison, and final item selection based on user instructions over noisy webpage observations. &
    \begin{tabular}[t]{@{}l@{}}
      Training: 1,824  \\
      Test: 210
    \end{tabular} \\

    \midrule

    \multicolumn{3}{c}{\textbf{StableToolBench}} \\
    \midrule
    Multi-turn tool-use tasks with executable tool interactions &
    Tool invocation, intermediate execution feedback handling, and solvability-aware evaluation under realistic function-calling constraints. &
    \begin{tabular}[t]{@{}l@{}}
      Training: 1,972 \\
      Test: 169
    \end{tabular} \\
    \bottomrule
  \end{tabularx}
  \caption{Dataset statistics. We report dataset splits following the standard benchmark protocol.}
  \label{tab:benchmark_stats}
\end{table}

\subsection{Data Preprocessing}

We use the provided expert trajectories as supervision to warm-start the agent policy. For the world model, we repurpose the same expert rollouts into transition-level supervision for a text world model by emitting one record per environment step, containing: (i) a compact state string (goal + current observation/page state/tool context + optional inventory or execution feedback), (ii) the executed expert action, and (iii) the next observation, optionally augmented with scalar signals such as \texttt{reward} and \texttt{done}. All records are stored in JSONL format and serialized as dialogue-style causal-LM input--output pairs, following standard SFT practice for text-based world models. The state serialization is benchmark-specific but follows a unified transition-level format. For ALFWorld and ScienceWorld, the state mainly consists of the task instruction, current observation, and optional inventory. For WebShop, the state additionally includes webpage content, product attributes, and browsing context. For StableToolBench, the state further incorporates tool descriptions, intermediate tool calls, and execution feedback. This unified serialization allows the same world-model training and adaptive-lookahead pipeline to be applied across embodied, web, and tool-use settings.

To annotate the number of lookahead steps (i.e., $K$) in \textbf{$\texttt{ITP}_{\text{R}}$}, we precompute imagined rollouts using the trained world model for a small discrete set of horizons (\texttt{k\_candidates}) at each expert step. 
Each imagined rollout is summarized into a short lookahead text snippet (\texttt{lookahead\_summary}). We then compute a per-horizon score that balances improved expert-action likelihood against deeper rollout cost, and store the resulting pseudo label (\texttt{k\_label}) along with all scores. This adaptive-horizon annotation procedure is applied consistently across all four benchmarks under their respective state/action serialization formats.

\section{Additional Baseline Comparisons}
\label{app:additional_comparisons}

\subsection{Tree Search and RL-from-World-Model Methods}
\label{app:additional_baselines}
\begin{table*}[t]
\centering
\small
\setlength{\tabcolsep}{6pt}
\begin{tabular}{lcc|lcc}
\toprule
\multicolumn{3}{c|}{\textbf{Tree Search-based}} & \multicolumn{3}{c}{\textbf{World-model-based Training}} \\
\cmidrule(r){1-3}\cmidrule(l){4-6}
\textbf{Method} & \textbf{SR (\%)} $\uparrow$ & \textbf{NB} $\downarrow$ &
\textbf{Method} & \textbf{SR (\%)} $\uparrow$ & \textbf{NB} $\downarrow$ \\
\midrule
RAP  & 28.57 & 0.42 & SFT   & 70.71 & --- \\
ToT  & 16.29 & 0.58 & D2PO  & 76.28 & 0.29 \\
MCTS & 25.00 & 0.39 & VaGen & 78.89 & 0.18 \\
\midrule
\textbf{$\texttt{ITP}_{\text{I}}$ (Ours)} & \textbf{41.43} & \textbf{0.25} &
\textbf{$\texttt{ITP}_{\text{R}}$ (Ours)} & \textbf{88.57} & \textbf{0.21} \\
\bottomrule
\end{tabular}
\caption{Comparison with additional baselines on ALFWorld. Left: tree-search-based methods under the same backbone/world-model interface. Right: RL-from-world-model baselines adapted to our text-only setting using the same state/action serialization and evaluation protocol. \texttt{ITP} consistently achieves the strongest overall performance, and for prompting-based methods also the best SR--budget trade-off.}
\label{tab:additional_baselines}
\end{table*}

\begin{table*}[t]
\centering
\small
\setlength{\tabcolsep}{7pt}
\begin{tabular}{lccc}
\toprule
\textbf{Method} & \textbf{World Model} & \textbf{Prompting-based SR}  & \textbf{Training-based SR}  \\
\midrule
Self-Refine            & \textcolor{red!70!black}{\xmark} & 22.59 & 68.89 \\
Multi-Attempt          & \textcolor{red!70!black}{\xmark} & 20.48 & 59.68 \\
K-candidate Selection  & \textcolor{red!70!black}{\xmark} & 36.42 & 78.14 \\
\midrule
\textbf{$\texttt{ITP}_{\text{I}}$} (Ours) & \textcolor{green!60!black}{\cmark} & $\mathbf{41.43}_{\textcolor{green!60!black}{\scriptscriptstyle +5.01}}$ & --- \\
\textbf{$\texttt{ITP}_{\text{R}}$} (Ours) & \textcolor{green!60!black}{\cmark} & --- & $\mathbf{88.57}_{\textcolor{green!60!black}{\scriptscriptstyle +10.43}}$ \\
\bottomrule
\end{tabular}
\caption{Compute-matched comparison with no-world-model alternatives. Subscripts denote the gain over the strongest no-world-model baseline in each setting.}
\label{tab:compute_matched}
\end{table*}

We additionally compare \model against two baseline families that represent natural alternatives to our formulation.
For inference-time planning, we include \textbf{ToT}~\citep{yao2023tree} and \textbf{MCTS}~\citep{kocsis2006bandit}, implemented under the same backbone/world-model interface and evaluated on ALFWorld with \textbf{Success Rate (SR)} and \textbf{Normalized Budget (NB)}.
For training-based comparison, we include \textbf{D2PO}~\citep{wang-etal-2025-world} and \textbf{VaGen}~\citep{wang2025vagen}.
Since D2PO and VaGen were originally developed for VLM settings, we adapt their training objectives to our text-only environment using the same state/action serialization, benchmark interface, and evaluation protocol as our method.
All methods use the same executable action space and are compared under matched backbone and benchmark settings.

Table~\ref{tab:additional_baselines} reports the results.
We obtain the following findings:
(1) \textbf{Tree search alone is insufficient}. Both ToT and MCTS underperform $\texttt{ITP}_{\text{I}}$ and are less budget-efficient, while $\texttt{ITP}_{\text{I}}$ achieves the best trade-off (41.43 SR / 0.25 NB versus RAP at 28.57 / 0.42).
(2) \textbf{RL-from-world-model training is helpful but not sufficient}. D2PO and VaGen consistently improve over Base SFT (70.71 $\rightarrow$ 76.28/78.89), but $\texttt{ITP}_{\text{R}}$ further advances the frontier to 88.57 SR with only a modest NB increase (0.21).
Taken together, these results suggest that the key benefit of our framework lies not in search or world-model usage alone, but in integrating world-model-based learning with \emph{adaptive lookahead} as an explicit budget/error control mechanism.

\subsection{Compute-Matched Comparisons with No-World-Model Alternatives}
\label{app:compute_matched}

To test whether \textbf{$\texttt{ITP}$} gains are merely due to extra compute, we compare against \emph{compute-matched} no-world-model baselines: \textbf{Self-Refine}, \textbf{Multi-Attempt}, and \textbf{K-candidate Selection}. These methods spend a similar budget on repeated direct action generation or refinement, but do not perform future-state imagination.

Table~\ref{tab:compute_matched} shows that no-world-model alternatives still lag behind \textbf{$\texttt{ITP}$} under matched compute. \textbf{$\texttt{ITP}_{\text{I}}$} improves over the strongest prompting-based baseline by 5.01 points, and \textbf{$\texttt{ITP}_{\text{R}}$} improves over the strongest training-based baseline by 10.43 points. This suggests that the gains come not from extra compute alone, but from allocating it to \textbf{state-conditioned adaptive lookahead over imagined future states}.

\section{Implementation Details}
\label{appendix:implement}
For fine-tuning, we employed several open-source models, including
Qwen3-8B~\citep{qwen3}\footnote{\url{https://huggingface.co/Qwen/Qwen3-8B/blob/main/LICENSE}},
Qwen2.5-7B~\citep{qwen2.5}\footnote{\url{https://huggingface.co/Qwen/Qwen2.5-7B-Instruct/blob/main/LICENSE}},
and Llama3.1-8B~\citep{dubey2024llama}\footnote{\url{https://huggingface.co/meta-llama/Llama-3.1-8B-Instruct/blob/main/LICENSE}}.
All experiments were conducted on a computational cluster equipped with 2$\times$ NVIDIA A100 80GB GPUs.
We report additional implementation details for reproducibility, including world-model training details, deployment profiling, stage-wise training cost of $\texttt{ITP}_{\text{R}}$, and the prompting templates.

\subsection{World Model Training Details}
\label{app:wm_training_details}

Our text world model is trained to predict the next textual observation conditioned on the current textual state and executed action, i.e., to model $p_{\phi}(s_{t+1}\mid s_t, a_t)$. Following the transition serialization described in Appendix~\ref{app:data}, each training instance is formatted as a dialogue-style causal-LM example, where the input contains the task instruction, current observation, optional inventory or tool feedback, and action, and the target is the next observation.

\begin{figure}[t]
  \centering
  \includegraphics[width=0.9\linewidth]{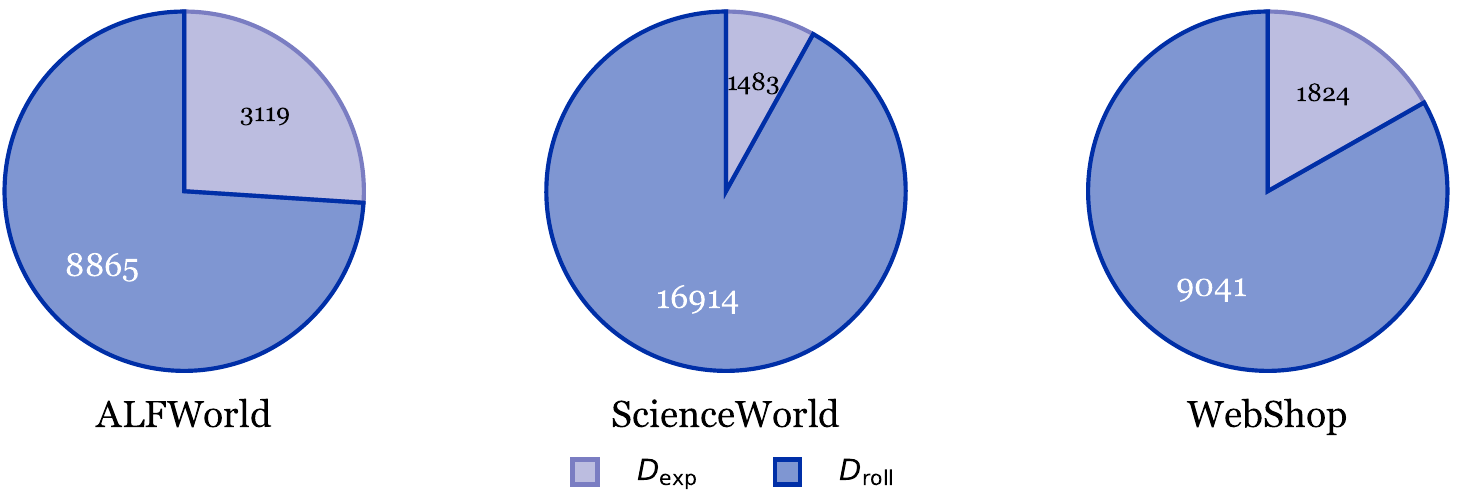}
  \caption{Visualization of world-model training data scale across benchmarks. For each benchmark, we show the number of expert episodes ($D_{\text{exp}}$) and rollout episodes ($D_{\text{roll}}$) used for training data construction.}
  \label{fig:wm_data_scale}
  \vspace{-8pt}
\end{figure}

We further construct adaptive-horizon supervision by scoring imagined rollouts over a small discrete set of horizon candidates at each expert step, producing one $K$-label record per step. Figure~\ref{fig:wm_data_scale} visualizes the resulting data scale across benchmarks, while Table~\ref{tab:wm_training_diagnostics} reports the optimization setup and convergence diagnostics for representative world-model training runs.

\begin{table}[t]
\centering
\small
\setlength{\tabcolsep}{6pt}
\renewcommand{\arraystretch}{1.08}
\begin{tabularx}{\columnwidth}{@{}lX@{}}
\toprule
\textbf{Item} & \textbf{Value} \\
\midrule
Objective & NLL / CE on $p_{\phi}(s_{t+1}\mid s_t, a_t)$ \\
Training epochs / steps & 3 epochs / 3,450 steps \\
Wall-clock time & 8.69 h \\
Throughput & 0.11 steps/s ; 1.764 samples/s \\
Held-out validation split & 5\% of $D_{\text{WM}}$, fixed seed \\
Best / final validation loss & 0.052 / 0.055 \\
Effective batch size (incl. accum.) & 16 \\
Max sequence length / truncation & 2048 \\
Optimizer + LR schedule & AdamW + cosine decay with warmup \\
Precision + hardware & bf16 on 2$\times$ NVIDIA A100 GPUs \\
Rollout collection cost & 9.5 GPU-hours on 1$\times$ A100 GPU \\
\bottomrule
\end{tabularx}
\caption{World-model training diagnostics for Qwen3-8B as the backbone. We report optimization settings, convergence indicators, and rollout collection cost.}
\label{tab:wm_training_diagnostics}
\end{table}

\subsection{Deployment Profiling}
\label{app:deployment_profiling}

Normalized Budget (NB) captures algorithmic token efficiency, but does not directly reflect deployment-facing runtime characteristics. We therefore additionally profile interactive inference and report wall-clock latency, throughput, and GPU memory usage under representative deployment settings.

\begin{table}[t]
\centering
\small
\setlength{\tabcolsep}{6pt}
\renewcommand{\arraystretch}{1.08}
\begin{tabularx}{\columnwidth}{@{}lccc@{}}
\toprule
\textbf{Metric} & \textbf{p50} & \textbf{p95} & \textbf{p99} \\
\midrule
Decision latency per step (s) & 14.2 & 31.8 & 44.7 \\
Policy time per step (s)      & 5.1  & 11.6 & 17.4 \\
Imagination time per step (s) & 7.8  & 18.9 & 28.5 \\
Tokens / second               & 43.5 & 31.2 & 24.0 \\
\midrule
\multicolumn{4}{l}{Episodes / hour: 11.8} \\
\multicolumn{4}{l}{Peak allocated / reserved VRAM (policy GPU): 31.0 / 38.4 GB} \\
\multicolumn{4}{l}{Peak allocated / reserved VRAM (WM GPU): 28.7 / 35.9 GB} \\
\bottomrule
\end{tabularx}
\caption{Deployment-oriented profiling of $\texttt{ITP}_{\text{R}}$. We report interactive latency, throughput, and peak GPU memory usage.}
\label{tab:deployment_profiling}
\end{table}

\subsection{\texorpdfstring{$\texttt{ITP}_{\text{R}}$}{ITP_R} Training Algorithm}
\label{app:itpr_algorithm}

Algorithm~\ref{alg:itpr_algorithm_appendix} presents the full training procedure of $\texttt{ITP}_{\text{R}}$, including pseudo-label generation for adaptive horizon supervision, warm-up training with lookahead labels, and online actor--critic optimization.

\begin{algorithm}[t]
\algrenewcommand\algorithmicrequire{\textbf{Input:}}
\algrenewcommand\algorithmicensure{\textbf{Output:}}
\caption{$\texttt{ITP}_{\text{R}}$: Reinforced Training with Adaptive Lookahead}
\label{alg:itpr_algorithm_appendix}
{\footnotesize 
\begin{algorithmic}[1]
\Require Dataset $\mathcal{D}=\{(s_t,a_t^*)\}$; Agent policy $\pi_{\theta_0}$; World model $\mathcal{M}_\phi$; Parameters $K_{\max}$, $\lambda_K$, $\eta$, $\alpha$, $\beta$.
\Ensure Agent policy $\pi_\theta$; Lookahead predictor $P_\theta$.

\Statex {\color{blue}// Stage 1: Pseudo-Labeling Lookahead Horizon}
\For{each episode in $\mathcal{D}$}
\State Cache $\{\hat \tau_t^{(k)}\}_{k=0}^{K_{\max}}$ via $\mathcal{M}_\phi$ \& teacher-forced $a_t^*$.
\For{each step $t$}
\State $S_t(k)\triangleq \log p_{\theta_0}(a_t^*\mid s_t,\hat \tau_t^{(k)})$.
\State $\tilde K_t \leftarrow \arg\max\big[S_t(k)-\lambda_K k\big]$.
\State Store $(s_t,a_t^*,\tilde K_t)$ in $\mathcal{D}_K$.
\EndFor
\EndFor

\Statex {\color{blue}// Stage 2: Warm-Up Training with Lookahead}
\For{each episode in $\mathcal{D}_K$}
\State Sample $(s_t,a_t^*,\tilde K_t)\!\sim\!\mathcal{D}_K$.
\State Update $\theta$ via Eq.~(\ref{eq:warmup}).
\EndFor

\Statex {\color{blue}// Stage 3: Online Actor--Critic Optimization}
\For{each episode in $\mathcal{D}$}
\State Sample $K_t\!\sim\!P_\theta(K\!\mid\!s_t)$, query $\mathcal{M}_\phi$ for $\hat \tau_t^{(K_t)}$.
\State Update $\theta$ via Eq.~(\ref{eq:a2c_loss}).
\EndFor

\State \Return $\pi_\theta$ and $P_\theta$.
\end{algorithmic}}
\end{algorithm}

\subsection{Training Cost of $\texttt{ITP}_{\text{R}}$}
\label{app:itr_training_cost}

To improve training-cost transparency, we report the stage-wise runtime of $\texttt{ITP}_{\text{R}}$. The training pipeline consists of three stages: (i) pseudo-label generation for adaptive horizon supervision, (ii) supervised warm-up of the policy and $K$-head, and (iii) online actor--critic optimization.

\begin{table}[t]
\centering
\resizebox{1.0\linewidth}{!}{
\begin{tabular}{lcc}
\toprule
\textbf{Stage} & \textbf{Wall-clock} & \textbf{GPU-hours} \\
\midrule
I: Lookahead Pseudo-Labeling    & 6.63 h  & 13.27 \\
II: Warm-Up Training   & 6.38 h  & 12.77 \\
III: Online Optimization  & 4.78 h  & 9.57 \\
\midrule
Total & 17.80 h & 35.60 \\
\bottomrule
\end{tabular}
}
\caption{Stage-wise training cost of $\texttt{ITP}_{\text{R}}$. Stage I corresponds to pseudo-label generation, Stage II to warm-up supervised training, and Stage III to online actor--critic optimization.}
\label{tab:itr_training_cost}
\end{table}

\subsection{Parameter Settings}

\begin{table}[th!]
\centering
\resizebox{0.96\linewidth}{!}{
\begin{tabular}{l l}
\toprule
\textbf{Name} & \textbf{Value} \\
\midrule
\multicolumn{2}{l}{\textit{ Warm-Up Training} } \\
\midrule
cutoff\_len & 2048 \\
epochs & 3 \\
per\_device\_train\_batch\_size & 1 \\
gradient\_accumulation\_steps & 16 \\
learning\_rate & $2\times 10^{-5}$ \\
warmup\_ratio & 0.03 \\
lr\_scheduler & cosine \\
fp16 / bf16 & False / Ture \\
gradient\_checkpointing & False \\
lora\_r & 8 \\
lora\_alpha & 16 \\
lora\_dropout & 0.05 \\
merge\_lora & True \\
\midrule
\multicolumn{2}{l}{\textit{ Online A2C Optimization}} \\
\midrule
$\gamma$ (discount) & 0.99 \\
rl\_learning\_rate & $5\times 10^{-6}$ \\
$\lambda_{K}$ (lookahead penalty) & 0.2 \\
$\lambda_{\text{step}}$ (step cost) & 0.01 \\
success\_bonus & 0.01 \\
invalid\_action\_penalty & -0.1 \\
$\eta$& 0.5 \\
$\alpha$ & 1.0 \\
$\beta$ & 0.01 \\
max\_grad\_norm & 1.0 \\
\midrule
\multicolumn{2}{l}{\textit{Inference Stage}} \\
\midrule
do\_sample (exploration) & True \\
temperature (exploration) & 0.7 \\
top\_p (exploration) & 0.9 \\
action\_max\_new\_tokens & 16 \\
imagine\_action\_max\_new\_tokens & 12 \\
wm\_max\_new\_tokens & 192 \\
$K_{\max}$ (ALFWorld) & 5 \\
$K_{\max}$ (ScienceWorld) & 8 \\
$K_{\max}$ (WebShop) & 7 \\
$K_{\max}$ (StableToolBench) & 3 \\
\bottomrule
\end{tabular}
}
\caption{Hyperparameter setup.}
\label{tab:hyperparams}
\end{table}

Table~\ref{tab:hyperparams} summarizes the hyperparameters used for training and inference. Unless otherwise specified, we apply the same configuration across different backbone models.
During exploration, the agent samples actions with temperature $0.7$.
Due to the difference in task trajectory lengths and planning horizons across benchmarks (e.g., ALFWorld has an average of 8 steps, ScienceWorld 15, WebShop 11, and StableToolBench 4), during \textbf{$\texttt{ITP}_{\text{R}}$} training, we set $K_{\max}$ to 5 for ALFWorld, 8 for ScienceWorld, 7 for WebShop, and 3 for StableToolBench.

\subsection{Comparison with Strong Planning and Search Baselines}
\label{app:strong_baselines}

We further compare \model with recent search-based, world-model-based, and
trajectory-optimization methods on ALFWorld. All methods use the same
benchmark split, executable action space, backbone family, and success-rate
metric as \model. 

\begin{table}[t]
\centering
\small
\begin{tabular}{lc}
\toprule
Method & ALFWorld SR (\%) $\uparrow$ \\
\midrule
CoT & 13.57 \\
ReAct & 19.29 \\
ToT & 16.29 \\
MCTS & 25.00 \\
RAP & 28.57 \\
\textbf{ITP$_I$} & \textbf{41.43} \\
\bottomrule
\end{tabular}
\caption{Comparison with training-free planning and search methods on
ALFWorld.}
\label{tab:training_free_baselines}
\end{table}

\begin{table}[t]
\centering
\small
\begin{tabular}{lc}
\toprule
Method & ALFWorld SR (\%) $\uparrow$ \\
\midrule
SFT & 70.71 \\
ETO & 70.50 \\
D2PO & 76.28 \\
VAGEN & 78.89 \\
WKM & 79.29 \\
QLASS & 80.35 \\
IWM & 82.14 \\
\textbf{ITP$_R$} & \textbf{88.57} \\
\bottomrule
\end{tabular}
\caption{Comparison with training-based planning methods on ALFWorld.}
\label{tab:training_based_baselines}
\end{table}

ITP$_I$ outperforms RAP, the strongest training-free baseline, by
12.86 percentage points. ITP$_R$ improves over IWM, the strongest
training-based baseline, by 6.43 points. These results show that the advantage
of ITP cannot be explained solely by search, additional rollout computation,
or world-model-based policy training. Instead, ITP benefits from adaptively
determining when imagination should be invoked and how deeply the model
should look ahead.

\subsection{Prompting Templates}
We provide the prompt templates for the agent policy and for the adaptive $K$-step lookahead inference procedure of \textbf{$\texttt{ITP}_{\text{I}}$}.
Figure~\ref{fig:prompt_react} shows the prompting template for the ReAct baseline.
Figure~\ref{fig:prompt_decide_k} illustrates the adaptive horizon selector, where the policy model maps the task instruction and the \emph{current} textual state $s_t$ to a single discrete lookahead depth $K\in[0,K_{\max}]$.
Figure~\ref{fig:prompt_imagine} presents the world-model foresight generator, which conditions only on $s_t$ and produces a concise $K$-step imagined trajectory enclosed by \texttt{<Foresight>...</Foresight>}.
Figure~\ref{fig:prompt_reflect_and_act} depicts the foresight-conditioned agent prompt, where the policy model consumes the task, the current state $s_t$, and the generated foresight to produce a ReAct-style response, and outputs an admissible environment action via exact copying.

\begin{figure*}[t]
    \centering
    \begin{tcolorbox}[
        title=\textbf{Prompt Template for Base Agent Policy},
        colframe=blue!50!black,
        colback=blue!5!white,
        coltitle=white,
        fonttitle=\bfseries,
        arc=1mm,
        boxsep=2pt,
        fontupper=\small
     ]
        Interact with a household to solve a task. You are an intelligent agent in a household environment.
        Your goal is to perform valid actions to complete the task goal.

        At each step $t$, you will be given the \textbf{task goal}, the \textbf{current state} (observation and optional inventory),
        and the \textbf{previous step context} (the previous action and its resulting environment feedback/observation).
        You must follow the \textbf{ReAct} paradigm: \textbf{Reason} $\rightarrow$ \textbf{Thought} $\rightarrow$ \textbf{Action}.

        \vspace{4pt}
        \textbf{You need to process the information in a specific order:}
        \begin{enumerate}[nosep, leftmargin=*]
            \item \textbf{Reason}: Briefly interpret the current state and the latest environment feedback. Identify what has been achieved and what remains.
            \item \textbf{Thought}: Plan the next few steps to make progress toward the task goal based on the reasoning.
            \item \textbf{Action}: Output \textbf{exactly one} next action that is valid in environment.
        \end{enumerate}

        After each turn, the environment will provide immediate feedback (a new observation, and optionally inventory updates).
        If the environment outputs ``Nothing happened'', the previous action is invalid; revise your plan and try a different valid action.

        \vspace{4pt}
        \textbf{Your response must use the following format:}
        \begin{tcolorbox}[colback=white, colframe=gray!30, arc=0mm, boxsep=0pt, left=4pt, right=4pt, top=2pt, bottom=2pt]
            \ttfamily\footnotesize
            Reason: <brief interpretation of state/feedback>\\
            Thought: <short plan for next steps>\\
            Action: <EXACTLY ONE valid action line>
        \end{tcolorbox}

        \vspace{4pt}
        \textbf{Inputs at step $t$:}
        \begin{tcolorbox}[colback=white, colframe=gray!30, arc=0mm, boxsep=0pt, left=4pt, right=4pt, top=2pt, bottom=2pt]
            \ttfamily\footnotesize
            Task goal:\par
            \{task\_goal\}\par\medskip
            Current state (observation + optional inventory):\par
            \{state\}\par\medskip
            Previous action (optional):\par
            \{prev\_action\}\par\medskip
            Latest environment feedback / observation:\par
            \{feedback\}
        \end{tcolorbox}
    \end{tcolorbox}
    \caption{Prompt template used for base agent policy on ALFWorld and ScienceWorld benchmarks.}
    \label{fig:prompt_react}
\end{figure*}

\begin{figure*}[t]
    \centering
    \begin{tcolorbox}[
        title=\textbf{Adaptive Selection of $K$ (\texttt{PolicyModel.decide\_k})},
        colframe=blue!50!black,
        colback=blue!5!white,
        coltitle=white,
        fonttitle=\bfseries,
        arc=1mm,
        boxsep=2pt,
        fontupper=\small
    ]
    \textbf{System prompt:}
    You are a planning assistant. Your job is to decide how many steps of look-ahead are needed right now. Given a task instruction and the dialogue/action history, output a single integer $K$ in the range [0, $K_{\max}$]

    \textbf{Task instruction:}
    {task}

    \textbf{History trajectory} (thoughts, actions, observations so far):
    {history}

    \textbf{Question:} Output a single integer $K$ in [0, $K_{\max}$].

    \vspace{4pt}
    \textbf{Your response should be:}
    \begin{tcblisting}{
        colback=white,
        colframe=gray!30,
        arc=0mm,
        boxsep=0pt,
        left=4pt,right=4pt,top=2pt,bottom=2pt,
        listing only,
        listing options={
            basicstyle=\ttfamily\footnotesize,
            breaklines=true,
            columns=fullflexible
        }
    }
K
    \end{tcblisting}
    \end{tcolorbox}

    \caption{Prompt used to adaptively select the lookahead horizon $K$ from the task instruction and trajectory history. The output must be a single integer within the range $[0,K_{\max}]$.}
    \label{fig:prompt_decide_k}
\end{figure*}

\begin{figure*}[t]
    \centering
    \begin{tcolorbox}[
        title=\textbf{$K$-Step Foresight Generation (\texttt{WorldModel.imagine})},
        colframe=blue!50!black,
        colback=blue!5!white,
        coltitle=white,
        fonttitle=\bfseries,
        arc=1mm,
        boxsep=2pt,
        fontupper=\small
    ]
    \textbf{System prompt:}
    You are a world model for the ALFWorld environment. Given an action/observation history, imagine the next few steps, describing likely observations and key objects.

    \vspace{4pt}
    \textbf{User prompt:}
    Predict the next {k} step(s). Return a concise plan inside <Foresight>...</Foresight> with numbered steps.
    \vspace{4pt}
    \textbf{User prompt:}
    History so far: {<history>}

    \vspace{4pt}
    \textbf{Your response should be:}
    \begin{tcblisting}{
        colback=white,
        colframe=gray!30,
        arc=0mm,
        boxsep=0pt,
        left=4pt,right=4pt,top=2pt,bottom=2pt,
        listing only,
        listing options={
            basicstyle=\ttfamily\footnotesize,
            breaklines=true,
            columns=fullflexible
        }
    }
Foresight state: <Foresight>...</Foresight>
    \end{tcblisting}
    \end{tcolorbox}

    \caption{Prompt used by the world model to generate a $K$-step foresight trajectory.}
    \label{fig:prompt_imagine}
\end{figure*}

\begin{figure*}[t]
    \centering
    \begin{tcolorbox}[
        title=\textbf{Foresight-Conditioned Action (\texttt{PolicyModel.reflect\_and\_act})},
        colframe=blue!50!black,
        colback=blue!5!white,
        coltitle=white,
        fonttitle=\bfseries,
        arc=1mm,
        boxsep=2pt,
        fontupper=\small
    ]
        You are an agent that \textbf{first imagines and then acts} .
        At each step, you will be given the task instruction, the \textbf{current state}, and a $K$-step foresight
        trajectory imagined by a world model. Use the foresight to reflect on progress and bottlenecks, then decide
        the next admissible action.

        \vspace{4pt}
        \textbf{You will be given:}
        \begin{enumerate}[nosep, leftmargin=*]
            \item \textbf{Task instruction} (\{task\})
            \item \textbf{Current state} (\{state\})
            \item \textbf{$K$-step foresight trajectory} from the world model (\{foresight\})
            \item \textbf{Admissible actions (instance-level)} (\{admissible\_actions\_joined\_by\_newlines\})
        \end{enumerate}

        After each turn, the environment will give you immediate feedback based on which you plan your next few steps.
        If the environment output ``Nothing happened'', that means the previous action is invalid and you should try more options.

        \vspace{4pt}
        \textbf{You must follow this order:}
        \begin{enumerate}[nosep, leftmargin=*]
            \item \textbf{Reflection}: Briefly assess whether the foresight indicates progress, contradictions, or missing subgoals.
            \item \textbf{Thought}: Provide a short plan for the next few steps based on the current state and foresight.
            \item \textbf{Action}: Output \textbf{exactly one} admissible action by \textbf{exact copying}.
        \end{enumerate}

        \vspace{4pt}
        \textbf{Hard constraint on the action:}
        You MUST choose the Action by copying \textbf{EXACTLY one line} from the provided \textbf{admissible actions} list.
        Do NOT paraphrase. Do NOT add extra tokens.

        \vspace{4pt}
        \textbf{User prompt:}
        \begin{tcolorbox}[
            colback=white,
            colframe=gray!30,
            arc=0mm,
            boxsep=0pt,
            left=4pt,right=4pt,top=2pt,bottom=2pt
        ]
            \ttfamily\footnotesize
            Task instruction:\par
            \{task\}\par\medskip
            Current state:\par
            \{state\}\par\medskip
            $K$-step foresight trajectory from the world model:\par
            \{foresight\}\par\medskip
            Admissible actions (copy exactly one line for Action):\par
            \{admissible\_actions\_joined\_by\_newlines\}
        \end{tcolorbox}

        \vspace{4pt}
        \textbf{Your response should use the following format:}
        \begin{tcolorbox}[
            colback=white,
            colframe=gray!30,
            arc=0mm,
            boxsep=0pt,
            left=4pt,right=4pt,top=2pt,bottom=2pt
        ]
            \ttfamily\footnotesize
            Reflection: <Reflection> ... </Reflection>\\
            Thought: <Thought> ... </Thought>\\
            Action: <Action> ... </Action>
        \end{tcolorbox}

    \end{tcolorbox}
    \caption{Prompt used to generate a \textbf{\model}-style response action conditioned on world-model foresight on ALFWorld and ScienceWorld benchmarks.}
    \label{fig:prompt_reflect_and_act}
\end{figure*}

\end{document}